\documentclass[twoside,11pt]{article}

\usepackage{jmlr2e}

\firstpageno{1}

\usepackage[utf8]{inputenc} 
\usepackage[T1]{fontenc}    
\usepackage{hyperref}       
\usepackage{url}            
\usepackage{booktabs}       
\usepackage{amsfonts}       
\usepackage{nicefrac}       
\usepackage{microtype}      
\usepackage[dvipsnames]{xcolor}
\usepackage{caption}

\usepackage{multirow}
\usepackage{color,comment,rotating,url,xspace} 
\usepackage{tikz}
\usepackage{subfig}
\usetikzlibrary{arrows,shadows,shapes,backgrounds,decorations,snakes,fit}
\usepackage[ruled,vlined]{algorithm2e}
\usetikzlibrary{shadows}
\usepackage{amsmath,amssymb,bm}
\usepackage{mathtools}
\usepackage{graphics}
\usepackage{wrapfig,lipsum}
\usepackage{paralist}
\newcommand*\circled[1]{\tikz[baseline=(char.base)]{
            \node[shape=circle,draw,inner sep=2pt] (char) {#1};}}
            
\makeatletter
\def\blfootnote{\xdef\@thefnmark{}\@footnotetext}
\makeatother

\begin{document}

\title{Icebreaker: Element-wise Active Information Acquisition \\with Bayesian Deep Latent Gaussian Model}

\author{\name Wenbo Gong$^{1*}$ \email wg242@cam.ac.uk \\
\name Sebastian Tschiatschek$^2$ \email Sebastian.Tschiatschek@microsoft.com \\
\name Richard Turner$^{12}$ \email ret26@cam.ac.uk\\
\name Sebastian Nowozin$^{2 \dagger}$ \email nowozin@gmail.com\\
\name Jos\'e Miguel Hern\'andez-Lobato$^{12}$ \email jmh233@cam.ac.uk\\
\name Cheng Zhang$^2$ \email Cheng.Zhang@microsoft.com\\
\addr $^1$Department of Engineering, University of Cambridge, Cambridge, UK\\
$^2$Microsoft Research, Cambridge, UK
}
\let\thefootnote\relax\footnote{$^*$Contributed during his internship in Microsoft Research\\
$^\dagger$Now at Google
AI, Berlin, Germany (contributed while being with Microsoft Research)\\Correspondence to: Cheng Zhang
and Wenbo Gong}

\maketitle

\begin{abstract}
In this paper we introduce the \emph{ice-start} problem, i.e., the challenge of {deploying} machine learning models when only little or no training data is initially available, and acquiring each feature element of data is associated with costs.
This setting is representative for the real-world machine learning applications. For instance, in the health-care domain,
when training an AI system for predicting patient metrics from lab tests, obtaining every single measurement comes with a high cost.
Active learning, where only the label is associated with a cost does not apply to such problem, because performing all possible lab tests to acquire a new training datum would be costly, as well as unnecessary due to redundancy.
We propose \textit{Icebreaker}, a principled framework to approach the ice-start problem.
\textit{Icebreaker} uses a full \textit{Bayesian Deep Latent Gaussian Model (BELGAM)} with a novel inference method. Our proposed method combines recent advances in amortized inference and stochastic gradient MCMC to enable fast and accurate posterior inference. By utilizing BELGAM's ability to fully quantify model uncertainty, we 
 also propose two information acquisition functions for \textit{imputation} and \textit{active prediction} problems.
We demonstrate that BELGAM performs significantly better than the previous VAE (Variational autoencoder) based models, when the data set size is small, using both machine learning benchmarks and real-world recommender systems and health-care applications. Moreover, based on BELGAM, Icebreaker further improves the performance and demonstrate the ability to use minimum amount of the training data to obtain the highest test time performance. 
\end{abstract}
\section{Introduction}
Acquiring information is costly in many real-world applications. For example,  to make a correct diagnosis and perform effective treatment, a medical doctor often needs to carry out a sequence of medical tests to gather information. Performing each of these tests is associated with a cost in terms of money, time, and health risks. 
To this end, an AI system should be able to suggest the information to be acquired in the form of "one measurement (feature) at a time" to enable the accurate predictions (diagnosis) for any new users/patients.
Recently, test-time active prediction methods, such as EDDI (Efficient Dynamic Discovery of high-value Inference) ~\citep{ma2018eddi},  provide a solution for when there is sufficient amount of training data available. 
Unfortunately, in these scenarios, training data can be also challenging and costly to obtain. For example, new data needs to be collected by taking measurements of currently hospitalized patients with their consent.
Ideally, we would like to deploy an AI system, such as EDDI when no or only limited training data is available.
We call this problem the \emph{ice-start} problem. 
It is desired to have a method to actively selected training data-element for such task.

The key to address the ice-start problem is to have a scalable model which
knows what it does not know, aka to quantify the epistemic uncertainty. In this way, this uncertainty can be used to guide the acquisition of training data, e.g., it would prefer to acquire unfamiliar but informative feature elements over others, such as the familiar but uninformative ones.
Thus, such an approach can reduce the cost of acquiring training data. We refer to this as element-wise training-time active acquisition.

Training-time active acquisition is needed in a great range of applications. Apart from general prediction tasks, such as the above health care example, general imputation tasks such as recommender system also need to address the ice-start problem. For example, when a new shop need a recommander where no historical customer data is available. Thus, a framework that can handle different type of tasks is desired.

Despite the success of element-wise test-time active prediction \citep{ma2018eddi,lewenberg2017knowing, shim2018joint,zannone2019odin}, few works have provided a general and scalable solutions for the problem of \emph{ice-start}. Additionally, these works \citep{melville2004active,krause2010utility,krumm2019traffic}  commonly are limited in a specific application scenario. 
An element-wise method needs to handle partial observations at any time. More importantly, we need to design new acquisition functions that takes the model parameter uncertainty into account.

In this work, we propose \textit{Icebreaker}, a principled framework to solve the ice-start problem. \textit{Icebreaker}  actively acquires informative feature elements during training and also perform active test prediction with small amount of data for training. To enable Icebreaker, we contribute:
\begin{compactitem}
    \item We propose a \underline{B}ayesian d\underline{e}ep \underline{L}atent \underline{Ga}ussian \underline{M}odel (BELGAM). Standard training of the deep generative model cares about the point estimates for the parameters, whereas our approach applies a fully Bayesian treatment to the weights. Thus, during the training time acquisition, we can leverage the epistemic uncertainty. (Section \ref{sec:BELGAM})
    \item We design a novel partial amortized inference method for BELGAM, naming PA-BELGAM. We combine  recent advances in amortized inference for the local latent variables and stochastic gradient MCMC for the model parameters, i.e. the weights of the neural network, to ensure  high inference accuracy. 
    (Section \ref{sec:PA_BELGAM}) 
    \item To complete Icebreaker, we propose two training-time information acquisition functions based on the uncertainties modeled by PA-BELGAM to identify informative elements. One acquisition function is designed for  imputation tasks (Section \ref{sec:impute}),  and the other for active prediction tasks. (Section \ref{sec:predict})
    \item We evaluate the proposed PA-BELGAM as well as the entire Icebreaker approach on well-used machine learning benchmarks and a real-world health-care task. The method demonstrates clear improvements compared to multiple baselines and shows that it can be effectively used to solve the ice-start problem. (Section \ref{sec:exp}) 
\end{compactitem}


\section{Bayesian Deep Latent Gaussian Model (BELGAM) with Partial Amortized Inference}
\label{sec:BELGAM}

As discussed before, to enable a generic and scalable solution for the Ice-start problem. We first need a flexible model with epidemic uncertainty quantification which can also handle missing values in the data. Here, we propose Bayesian Deep Latent Gaussian Model (BELGAM) with scaleable and accurate approximate inference.

\subsection{Bayesian Deep Latent Gaussian Model (BELGAM)}

\begin{wrapfigure}[10]{r}{0.3 \textwidth}
    \centering
\pgfdeclarelayer{background}
\pgfdeclarelayer{foreground}
\pgfsetlayers{background,main,foreground}

\begin{tikzpicture}

\tikzstyle{surround} = [thick,draw=black,rounded corners=1mm]
\tikzstyle{scalarnode} = [circle, draw, fill=white!11,  
    text width=1.2em, text badly centered, inner sep=2.5pt]
\tikzstyle{scalarnodenoline} = [fill=white!11, 
    text width=1.2em, text badly centered, inner sep=2.5pt]
\tikzstyle{arrowline} = [draw,color=black, -latex]
\tikzstyle{dashedarrowcurve} = [draw,color=black, dashed, out=100,in=250, -latex]
\tikzstyle{dashedarrowline} = [draw,color=black, dashed,  -latex]
    
    \node [scalarnode] at (1.5,0) (O)   {$\theta$};
    \node [scalarnode] at (0, 0) (Z) {\textbf{Z}};
    \node [scalarnode] at (0, -1.5) (X) {\textbf{X}};    
      \filldraw[black!80][] (X.south east)
           arc (-45:180-45:0.35)
           --cycle
          ;
    \node [scalarnode, fill opacity=0.7] at (0, -1.5) (X) {\textbf{X}};
    \node[surround, inner sep = .3cm] (f_N) [fit = (Z)(X) ] {};
    \path [arrowline] (Z) to (X);
    \path[arrowline]  (O) to (X);
\end{tikzpicture}
    \caption{BELGAM }
    \label{fig:graphicalModel}
\end{wrapfigure}

We design a flexible full Bayesian model which provides the model uncertainty quantification. A Bayesian latent variable generative model as shown in Figure \ref{fig:graphicalModel} is a common choice, but previous work of such models are typically linear and not flexible enough to model highly complex data. A Deep Latent Gaussian Model which uses a neural network mapping is flexible but not fully Bayesian as the uncertainty of the model itself is ignored.  We thus propose a \underline{B}ayesian D\underline{e}ep \underline{L}atent \underline{Ga}ussian \underline{M}odel (BELGAM),  which uses a Bayesian neural network to generate observations $\bm{X}_O$ from local latent variables $Z$ with global latent weights $\theta$ shown in \ref{fig:graphicalModel}. 
The model is thus defined as:
\begin{equation}
    p(\bm{X}_O,\theta,\bm{Z})=p(\theta)\prod_{i=1}^{|O|}\prod_{d\in O_i}p(x_{i,d}|\bm{z}_i,\theta)p(\bm{z}_i),
\end{equation}
where $|X_O|$ is the observed data.
 The goal is to infer the posterior, $p(\theta,\bm{Z}|\bm{X}_O)$, for both local latent variable $\bm{Z}=[\bm{z}_1,\ldots,\bm{z}_{|o|}]$ and global latent weights $\theta$. Given the posterior, we can infer the missing data as in \citep{ma2018eddi}.
 Such a model is generally intractable and approximate inference is needed  \citep{zhang2017advances,li2018approximate}. Variational inference (VI) \citep{wainwright2008graphical,zhang2017advances,jordan1999introduction,beal2003variational,li2018approximate} and sampling-based methods \citep{andrieu2003introduction} are two types of approaches used for this task. Sampling-based approaches are known for accurate inference performances and theoretical guarantees. 

However, sampling the local latent variable $\bm{Z}$ is computationally expensive as the cost scales linearly with the data set size. To best trade off the computational cost against the inference accuracy, we propose to amortize the inference for $\bm{Z}$ and keep an accurate sampling-based approach for the global latent weights $\theta$.  Specifically, we use preconditioned stochastic gradient Hamiltonian Monte Carlo  (SGHMC) \citep{chen2016bridging} (see appendix for details).  

\subsection{Partial Amortized BELGAM}
\label{sec:PA_BELGAM}

\paragraph{Revisiting amortized inference in the presence of missing data.}
Amortized inference \citep{kingma2014auto,Rezende2014StochasticBA} is an efficient extension for variational inference. It was originally proposed for deep latent Gaussian models where only local latent variables $\bm{Z}$ need to be inferred. Instead of using an individually parametrized approximation $q(\bm{z}_i)$ for each data instance $\bm{x}_i$, amortized inference uses a deep neural network as a function estimator to compute $q(\bm{z}_i)$ using $\bm{x}_i$ as input, $q(\bm{z}_i |x_i)$. Thus, the estimation of the local latent variable does not scale with data set size during model training. 

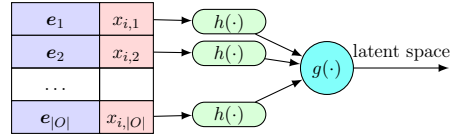
\begin{wrapfigure}[9]{r}{0.4 \textwidth}
    \centering
    \vspace{-10pt}
\pgfdeclarelayer{background}
\pgfdeclarelayer{foreground}
\pgfsetlayers{background,main,foreground}

\definecolor{babyblue}{rgb}{0.54, 0.81, 0.94}
\definecolor{bisque}{rgb}{1.0, 0.89, 0.77}
\definecolor{bittersweet}{rgb}{1.0, 0.44, 0.37}
\definecolor{pink}{rgb}{1,0.75,0.8}
\definecolor{cyan}{rgb}{0.019, 1, 0.972}
\usetikzlibrary{matrix}
\begin{tikzpicture}[cell/.style={rectangle,draw=black},
space/.style={minimum height=0.3cm,matrix of nodes,row sep=-\pgflinewidth,column sep=-\pgflinewidth},text depth=0.5ex,text height=2ex,nodes in empty cells,scale=0.65, every node/.style={scale=0.65}]
\tikzstyle{arrowline} = [draw,color=black, -latex]

\matrix (first) [space, column 1/.style={nodes={fill=blue!15,draw},minimum width=1.8cm},column 2/.style={nodes={fill=red!15,draw},minimum width=1.1cm},
row 3/.style={nodes={fill=none,draw}}
]
{
\node {$\bm{e}_1$}; & \node (r1) {$x_{i,1}$}; \\ 
\node {$\bm{e}_2$}; & \node (r2) {$x_{i,2}$}; \\
\node {$\ldots$}; & \node (r3) {};\\
\node {$\bm{e}_{|O|}$}; & \node (r4) {$x_{i,|O|}$}; \\
};
\node (rect) at (3,0.95) [draw, rounded rectangle,inner sep=0pt,fill=green!15,minimum width=1.5cm, minimum height=0.5cm] (h1){$h(\cdot)$};
\node (rect) at (3,0.3) [draw, rounded rectangle,inner sep=0pt,fill=green!15,minimum width=1.5cm, minimum height=0.5cm] (h2){$h(\cdot)$};
\node (rect) at (3,-0.95) [draw, rounded rectangle,inner sep=0pt,fill=green!15,minimum width=1.5cm, minimum height=0.5cm] (h4){$h(\cdot)$};

\path [arrowline]  (r1) to (h1);
\path [arrowline]  (r2) to (h2);
\path [arrowline]  (r4) to (h4);

\node [circle] at (5,0) [draw, minimum width=0.1cm, minimum height=0.1cm, fill=cyan!50] (g){$g(\cdot)$};
\path [arrowline]  (h1) to (g);
\path [arrowline]  (h2) to (g);
\path [arrowline]  (h4) to (g);
\draw [arrowline] (g) -- node[anchor=south] {latent space} (7.5,0);

\end{tikzpicture}
    \caption{
    The illustration of P-VAE inference network structure. 
  }
    \label{fig:PVAE}
\end{wrapfigure}

However, in our problem setting, the feature values for each data instance are partially observed. Thus, the vanilla amortized inference cannot be used as the input dimensionality to the network can vary for each data instance. As with the Partial VAE proposed in \cite{ma2018eddi}, we adopt the set encoding structure \citep{qi2017pointnet,zaheer2017deep} to build the inference network to infer $\bm{Z}$ based on partial observations in an amortized manner.

As shown in Figure \ref{fig:PVAE}, 
for each data instance $\bm{x}_i\in \bm{X}_O$ with $|o_i|$ observed features, the input is modified as $\bm{S}_i=[\bm{s}_{i,1},\ldots,\bm{s}_{i,|o_i|}]$ where $\bm{s}_{i,d}=[x_{i,d},\bm{e}_d]$ and $\bm{e}_d$ is the feature embedding. This is fed into a standard neural network $h(\cdot):\mathbb{R}^{M+1}\rightarrow \mathbb{R}^{K}$ where $M$ and $K$ are the dimensions of the latent space and $\bm{e}_{d}$ respectively. Finally, a permutation invariant set function $g(\cdot)$ is applied.  
In this way, we build an inference network structure that is compatible with partial observations.

\paragraph{Amortized inference + SGHMC}
As discussed previously, we want to be computationally efficient when inferring $\bm{Z}$ and be accurate when inferring the global latent weights $\theta$ for BELGAM. Thus, we start with VI and then nest SGHMC into it. Assume we have the factorized approximated posterior $q(\theta,\bm{Z}|\bm{X}_O)\approx q(\theta|\bm{X}_O)q_\phi(\bm{Z}|\bm{X}_O)$ \citep{kingma2014auto,ma2018eddi}, then the proposed inference scheme can be summarized into two stages: (i) Sample $\theta\sim q(\theta|\bm{X}_O)$ using SGHMC, (ii) Update the amortized inference network $q_\phi(\bm{z}_i|\bm{x}_i)$ to approximate $p(\bm{z}_i|\bm{x}_i)$. 

First, we present how to sample $\theta\sim q(\theta|\bm{X}_O)$ using SGHMC. The optimal form for $q(\theta|\bm{X}_O)$ can be defined as $q(\theta|\bm{X}_O)=\frac{1}{C}e^{\log p(\bm{X}_O,\theta)}$, where $C$ is the normalization constant $p(\bm{X}_O)$. In order to sample from such optimal distribution, the key is to compute the gradient $\nabla_{\theta}\log p(\bm{X}_O,\theta)$. Unfortunately, this is intractable due to marginalizing the latent variable $\bm{Z}$. Instead, we propose to approximate this quantity by transforming the marginalization into an optimization:
\begin{equation}
    \log p(\bm{X}_O,\theta)\geq \sum_{i\in \bm{X}_O}{\left[\mathbb{E}_{q_\phi(\bm{z}_i|\bm{x}_i)}[\log p(\bm{x}_i|\bm{z}_i,\theta)]-KL[q_\phi(\bm{z}_i|\bm{x}_i)||p(\bm{z}_i)]\right]}+\log p(\theta).
    \label{eq: Joint approximation}
\end{equation}
We call the right hand side of Eq. \ref{eq: Joint approximation} as $\mathcal{L}_{joint}(\bm{X}_O;\phi)$. Therefore, the marginalization of $\bm{Z}$ is transformed into an optimization problem
\begin{equation}
    \nabla_{\theta}\log p(\bm{X}_O,\theta)=\nabla_{\theta}\max_{q_\phi\in \mathcal{F}}{\sum_{i\in \bm{X}_O}{\left[\mathbb{E}_{q_\phi(\bm{z}_i|\bm{x}_i)}[\log p(\bm{x}_i|\bm{z}_i,\theta)]-KL[q_\phi(\bm{z}_i|\bm{x}_i)||p(\bm{z}_i)]\right]}+\log p(\theta)}.
    \label{eq: SGHMC ELBO}
\end{equation}
where $\mathcal{F}$ is a sufficiently large function class. 

After sampling $\theta$, we update the inference network with these samples by optimizing: 
\begin{equation}
\begin{split}
     &\mathcal{L}(\bm{X}_O;\phi)=\mathbb{E}_{q(\theta,\bm{Z}|\bm{X}_O)}[\log p(\bm{X}_O|\bm{Z},\theta)]-KL[q(\bm{Z},\theta|\bm{X}_O)||p(\bm{Z},\theta)]\\
     &=\mathbb{E}_{q(\theta|\bm{X}_O)}\left[\sum_{i\in\bm{X}_O}{\mathbb{E}_{q_\phi(\bm{z}_i|\bm{x}_i)}[\log p(\bm{x}_i|\bm{z}_i,\theta)]-KL[q_\phi(\bm{z}_i|\bm{x}_i)||p(\bm{z}_i)]}\right]-KL[q(\theta|\bm{X}_O)||p(\theta)].
\end{split}
\label{eq: Inference net ELBO}
\end{equation}
where the outer expectation can be approximated by SGHMC samples.  
The resulting inference algorithm resembles an iterative update procedure, like \textit{Monte Carlo Expectation Maximization} (MCEM) \citep{wei1990monte} where it samples latent $\bm{Z}$ and optimizes $\theta$ instead. We call the proposed model \textit{Partial Amortized BELGAM} (\textit{PA-BELGAM}). Partial VAE is actually a special case of PA-BELGAM, where $\theta$ is a point estimate instead of a set of samples.

Note that, in this way, the computational cost with single chain SGHMC is exactly the same as training a normal VAE thanks to the amortization for $\bm{Z}$. Thus, PA-BELGAM scales to large data when needed.
The only additional cost is the memory for storing $\theta$ samples. Thus, we adopt a similar idea based on the Moving Window MCEM algorithm \citep{havasi2018inference}, where samples are stored and updated in a fixed size pool with a first in first out (FIFO) procedure. 
In the next two sections, we present Icebreaker which utilize BELGAM for two general machine learning tasks separately: imputation tasks and prediction tasks.
\section{Icebreaker for Imputation Tasks}
\label{sec:impute}
We present Icebreaker for the imputation task first as the PA-BELGAM can be directly applied, such as recommander in the same way as \citet{ma2018partial}. We introduce the problem formulation first which provide an overview of the the method. We then present our proposed the active training acquisition function in detail.

\subsection{Problem Definition}
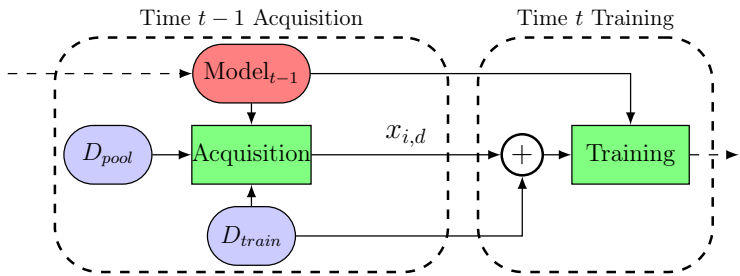
\begin{figure}[b]
\centering
\scalebox{1.2}{
\pgfdeclarelayer{background}
\pgfdeclarelayer{foreground}
\pgfsetlayers{background,main,foreground}

\definecolor{babyblue}{rgb}{0.54, 0.81, 0.94}
\definecolor{bisque}{rgb}{1.0, 0.89, 0.77}
\definecolor{bittersweet}{rgb}{1.0, 0.44, 0.37}
\definecolor{pink}{rgb}{1,0.75,0.8}
\definecolor{cyan}{rgb}{0.019, 1, 0.972}
\begin{tikzpicture}[scale=0.6,every node/.style={scale=0.65}]

\tikzstyle{round rect}= [draw, rounded rectangle,inner sep=0pt,minimum width=1.5cm, minimum height=1cm]
\tikzstyle{recta} = [draw, rectangle, inner sep=0pt, fill=green!50, minimum width=2cm, minimum height=1cm ]
\tikzstyle{arrowline} = [draw,color=black, -latex]
\tikzstyle{dash rect}=[draw,rectangle,rounded corners=5mm,thick,dashed]

\node (rect) at (-1.,0) [dash rect,minimum width=6.7cm, minimum height=4.cm] {};
\node (rect) at (5.35,0) [dash rect,minimum width=4cm, minimum height=4.cm] {};

\node (rect) at (-0.3-6.7/2,0) [round rect,fill=blue!20] (D Pool){\large $D_{pool}$};
\node (rect) at (-1,0) [recta] (Acquisition){\large Acquisition};
\node (rect) at (-1,1.5) [round rect, fill=red!50,minimum width=2cm] (Model){\large $\text{Model}_{t-1}$};
\node (rect) at (-1,-1.5) [round rect,fill=blue!20] (D Train){\large $D_{train}$}; 
\path [arrowline] (D Pool) to (Acquisition);
\path [arrowline] (Model) to (Acquisition);
\path [arrowline] (D Train) to (Acquisition);

\node [circle] at (4,0) [draw,thick, fill=none, minimum width=0.7cm, minimum height=0.7cm,inner sep=0pt] (plus){\Large $+$};

\path [arrowline] (Acquisition) to node[anchor=south]{\Large $x_{i,d}$} (plus);
\draw [arrowline] (D Train) -| (plus);
\node (rect) at (6.,0) [recta] (Training){\large Training};
\path [arrowline] (plus) to (Training);
\path [arrowline] (Model) -| (Training);
\path [arrowline,dashed] (-5.5,1.5) to (Model); 
\path [arrowline,dashed] (Training) to (8,0);

\node [] at (-1,2.5) {Time $t-1$ Acquisition};
\node [] at (5.35,2.5) {Time $t$ Training};
\end{tikzpicture}
}
\caption{The flow diagram of active training phase at time $t-1$. 
}
\label{fig:active training t-1}
\end{figure}

Assume at each training acquisition step we have training data $\mathcal{D}_{train}$, a pool data set $\mathcal{D}_{pool}$ that contain the data we could query and $\mathcal{D}_{train}\cup\mathcal{D}_{pool}= \bm{X}\in \mathbb{R}^{N\times D}$. In the ice-start scenario, $\mathcal{D}_{train} = \emptyset$. At each step of the training-time acquisition, we actively select data points $x_{i,d}\in\mathcal{D}_{pool}$ to acquire, thereby moving them into $\mathcal{D}_{train}$ and updating the model with the newly formed $\mathcal{D}_{train}$. Figure \ref{fig:active training t-1} shows the flow diagram of this procedure. During the process, at any step, there is an observed data set $\bm{X}_O$ (e.g. for training data set $\bm{X}_O=\mathcal{D}_{train}$) and unobserved set $\bm{X}_U$ with $|O|$ and $|U|$ number of rows respectively. For each (partially observed) data instance $\bm{x}_i\in \bm{X}_O$, we have the observed index set $O_i$ containing index of observed features for row $i$. The training time acquisition procedure is summarised in algorithm \ref{alg: training time acquisition} and Figure \ref{fig:active training t-1}.

\begin{algorithm}[H]
\SetAlgoLined
\SetNoFillComment
\SetKwInOut{Input}{input}

\Input{$\bm{X}_O$,$\bm{X}_U$,$\Phi$,$\mathcal{M}$, Acquisition number $K$}
\tcc{Initialization}
 $\bm{X}_O=\emptyset$\;
 \While{$\bm{X}_U\neq\emptyset$}{
  \tcc{Information acquisition}
  Compute reward $R(x_{i,d},\bm{X}_O)$ for $x_{i,d}\in\bm{X}_U$ using Eq.\ref{eq: Acquisition Imputation} or \ref{eq: Acquisition Combined} \tcp*{Reward computation}
  Sample $\bm{X}_{new}$ \tcp*{Sample $K$ feature elements according to the $R$ value.}
$\bm{X}_O= \bm{X}_O \cup \bm{X}_{new} $
  \tcp*{Update training set}
  \tcc{Model Training}
  Re-initialized the model $\mathcal{M}$ \tcp*{Re-initialization to avoid local optimum }
  $\mathcal{M}=$Train($\mathcal{M}$,$\Xi$)\;
  \tcc{Test task}
  Test($\mathcal{M}$)\tcp*{Test performance of the current model $\mathcal{M}$}
  
 }
 \caption{Element-wise training time acquisition}
 \label{alg: training time acquisition}
\end{algorithm}

\begin{algorithm}[H]
\SetAlgoLined
\SetNoFillComment
\SetKwInOut{Input}{input}
\KwResult{Evaluation metric $p$}
\Input{$\mathcal{D}_O$,$\mathcal{D}_U$,$\mathcal{M}$,$f(\cdot)$}
$\tilde{\mathcal{D}}_U=$Impute($\mathcal{M}$,$\mathcal{D}_O$)\tcp*{Imputation}
Compute $p=f({\mathcal{D}}_U)$
 \caption{Imputation task evaluation}
 \label{alg: Imputation}
\end{algorithm}

For training acquisition notations, we denote the training set $\mathcal{D}_{train}=\bm{X}_O$ and the pool set $\mathcal{D}_{pool}=\bm{X}_U$. The model $\mathcal{M}$ training hyper-parameters are grouped as $\Xi$. We also have the selection method Update($\bm{X}_O$,$\bm{X}_U$,$K$) which picks $K$ values from {the} pool into {the} training set using the acquisition described in below. The evaluation methods for the test task is denoted as Test($\mathcal{M}$), which is algorithm \ref{alg: Imputation}. $D_O$ and $D_U$ represents the observed/unobserved test data. $f(\cdot)$ is the evaluation metric e.g. negative log likelihood (NLL).
\subsection{Active Information Acquisition for Imputation}
\label{sec:Acquisition}
Designing the training time acquisition function is not trivial at all. Some of the previous work focus on developing intuition-based acquisition functions such as variance of the feature value \citep{huang2018active} or expected improvement of the model \citep{melville2005expected}. On the other hand, the recent proposed acquisition, EDDI \citep{ma2018eddi} is  an valid information-theoretic objective. However, it is under the assumptions that the model is well trained and only test-time feature acqusition is performed, thus, this is not directly applicable in our formulation (see appendix \ref{subsec: EDDI} for details). The ideal acquisition should balance the reduction of the model uncertainty and performance of the desired task.

Imputing missing values is important to applications such as recommender systems and other down-stream tasks.
In this setting, the goal is to learn about all the features elements (item user pairs in the recommender system setting) as quickly as possible. This can be formalized as selecting the elements $x_{i,d}$ that maximize the expected reduction in the posterior uncertainty of $\theta$:
\begin{equation}
    R_{I}(x_{i,d},\bm{X}_O)=H[p(\theta|\bm{X}_O)]-\mathbb{E}_{p(x_{i,d}|\bm{X}_O)}[H[p(\theta|\bm{X}_O,x_{i,d})]].
    \label{eq: Acquisition Imputation}
\end{equation}
We use the symmetry of mutual information to sidestep the posterior update $p(\theta|\bm{X}_O,x_{i,d})$ and entropy estimation of $\theta$ for efficiency. Thus, Eq. \ref{eq: Acquisition Imputation} is written as 
\begin{equation}
     R_{I}(x_{i,d},\bm{X}_O)= H[p(x_{i,d}|\bm{X}_O)]-\mathbb{E}_{p(\theta|\bm{X}_O)}[H[p(x_{i,d}|\theta,\bm{X}_O)]].
    \label{eq: True Acquisition Imputation}
\end{equation}
We can approximate Eq. \ref{eq: True Acquisition Imputation} as 
\begin{equation}
        R_{I}(x_{i,d},\bm{X}_O)\approx -\frac{1}{K}\sum_{k}{\log \frac{1}{MN} \sum_{m,n}{p(x_{i,d}^k|\bm{z}_{i}^m,\theta^n)}}+\frac{1}{NK}\sum_{k,n}\log \frac{1}{M}\sum_{m}{p(x_{i,d}^k|\bm{z}_i^m,\theta^n)}.
        \label{eq: Acquisition Imputation approximation}
\end{equation}
based on the samples $\{\theta^n\}_{n=1}^N$, $\{\bm{z}_i^m\}_{m=1}^M$ and $\{x_{i,d}^k\}_{k=1}^K$ from SGHMC, the amortized inference network and the data distribution respectively. The sample $x_{i,d}\sim p(x_{i,d}|\bm{X}_O)$ can be generated in the following way: (i) $\bm{z}_i\sim q_\phi(\bm{z}_i|\bm{x}_{io})$ (ii) $\theta\sim q(\theta|\bm{X}_O)$ (iii) $x_{i,d}\sim p(x_{i,d}|\theta,\bm{z}_i)$, where $\bm{x}_{io}$ represents the observed features in $i^{th}$ row of $\bm{X}_O$

\section{Icebreaker for Prediction Tasks}
\label{sec:predict}
Next, we introduce the second type of test task called active prediction, where a target variable is specified and active sequential acquisition is used in the test time. Note that the same framework applies for regular prediction tasks as well. Here, we demonstrate the case where feature wise active information acquisition is used in both training and testing time, which is desired in the data costly situation. Following the same structure of imputation section, we first formally define this problem, and then extend the model used for imputation to active prediction. In the end, a novel acquisition function is proposed.

\subsection{Problem Definition}
During the training acquisition, the procedure is exactly the same as imputation task, which is shown in algorithm \ref{alg: training time acquisition} and Figure \ref{fig:active training t-1}, apart from that we have clear target variables. We denote the target as $\bm{Y}$. In this case, each $\bm{x}_i\in\bm{X}_O$ has a corresponding target $\bm{y}_i$. In addition, for each training acquisition step, instead of querying single feature value $x_{i,d}$ as the imputation task, we query a feature-target pair $(x_{i,d},y_i)$ if $y_i$ has not been queried before. Otherwise, we only query $x_{i,d}$. 

As the name active prediction suggests, the test task requires the model to actively select features in test pool set to improve the prediction accuracy. Formally, we have an additional test target $\bm{Y}^*$ compared to imputation task. At each test time query, a single feature $x^*_{id}\in \bm{X}^*_U$ for each row $i$ is moved into $\bm{X}^*_O$. The goal is to achieve better target prediction $f({\bm{Y}}^*)$ with minimal data queries in the test time. We use \textit{Area under information curve} (AUIC) as the evaluation metric for this test-time active prediction suggested in EDDI \citep{ma2018eddi}. The entire active prediction evaluation procedure is summarised in algorithm \ref{alg: active prediction}
\begin{algorithm}[H]
\SetAlgoLined
\SetNoFillComment
\SetKwInOut{Input}{input}
\KwResult{Area under information curve AUIC }
\Input{$\mathcal{D}_O$,$\mathcal{D}_U$,$\mathcal{M}$,$f(\cdot)$,$\bm{Y}$}
\tcc{Initialization}
 $\mathcal{D}_O=\emptyset$\;
 AUIC$=0$\;
 \While{$\mathcal{D}_U\neq\emptyset$}{
  \tcc{Test time acquisition}
  Compute EDDI reward $R(x_{i,d},\mathcal{D}_O)$ for $x_{i,d}\in\mathcal{D}_U$ using Eq.\ref{eq: EDDI Computable} for each row $i$\;
  Select single $x_{i,d}\in\mathcal{D}_U$ into $\mathcal{D}_O$ for each row $i$ \tcp*{Test time acquisition}
  \tcc{Test Evaluation}
  Predict $\tilde{\bm{Y}}=$Predict($\mathcal{M}$,$\mathcal{D}_O$)\tcp*{Prediction}
  Compute $p=f(\bm{Y})$ \tcp*{Evaluation}
  AUIC+=$p$\tcp*{Compute AUIC value}

 }
 \caption{Active prediction task evaluation}
 \label{alg: active prediction}
\end{algorithm}.

\subsection{Model and Active Information Acquisition for Active Prediction}
\paragraph{Conditional BELGAM}
In the prediction task, the model needs to incorporate the target variable.
The proposed model and inference algorithm in the previous imputation section can be easily extended to incorporate these variables. In general, PA-BELGAM can be adapted in any VAE based framework directly. One possible choice is to adopt the formulation of conditional VAE \citep{sohn2015learning}. The details are in appendix \ref{section: Conditional BELGAM}.

\paragraph{Icebreaker for active target prediction.}
For the prediction task, solely reducing the model uncertainty is not optimal as the goal is to predict the target variable $\bm{Y}$. In this context, we
require the model to (1) capture the correlations and accurately impute the unobserved feature values in the pool set because during the test time sequential feature selection, the model needs to estimate the candidate missing element $x_{i,d}$ for decision making, and (2) find informative feature combinations to learn to predict the target variable. 
Thus, the desired acquisition function needs to trade-off exploring different features to learn their relationships against learning a predictor by exploiting the informative feature combinations.
We propose the following objective: 
\begin{equation}
    R_{P}(x_{i,d},\bm{X}_O)=\mathbb{E}_{p(x_{i,d}|\bm{X}_O)}[H[p(\bm{y}_i|x_{i,d},\bm{X}_O)]]-\mathbb{E}_{p(\theta,x_{i,d}|\bm{X}_O)}[H[p(\bm{y}_i|\theta,x_{i,d},\bm{X}_O)]].
    \label{eq: Acquisition Function Prediction}
\end{equation}
The above objective is equivalent to \textit{conditional mutual information} $I(\bm{y}_i,\theta|x_{i,d};\bm{X}_O)$. Thus, maximizing it is the same as maximizing the information to predict the target $y_i$ through the model weights $\theta$, conditioned on the observed features $X_O$ with this additional feature $x_{i,d}$. In our case, the $x_{i,d}$ is unobserved. 
As the weights $\theta$ do not change significantly over one feature element, for computational convenience we assume $p(\theta|\bm{X}_O)\approx p(\theta|\bm{X}_O,x_{i,d})$ when estimating the objective. 

\noindent Similar to Eq. \ref{eq: Acquisition Imputation approximation}, we approximate this objective using Monte Carlo integration:
\begin{equation}
\begin{split}
     &R_P(x_{i,d},\bm{X}_O)\approx\\ &-\frac{1}{JK}\sum_{j,k}{\log \frac{1}{MN}\sum_{m,n}{p(\bm{y}_i^{(j,k)}|\bm{z}_i^{(m,k)},\theta^n)}}+\frac{1}{KNJ}\sum_{j,n,k}{\log \frac{1}{M}\sum_{m}{p(\bm{y}_i^{(j,k)}|\bm{z}_i^{(m,k)},\theta^n)}},
\end{split}
    \label{eq: Acquisition Prediction Approximation}
\end{equation}
where we draw
$\{\bm{z}_i^{(m,k)}\}_{m=1}^M$ from $q_{\phi}(\bm{z}_i|\bm{X}_O,x_{i,d}^k)$ for each imputed sample $x_{i,d}^k$. Others ($\{\theta^n\}_{n=1}^N$, $\{\bm{y}_i^{(j,k)}\}_{j=1}^J$ and $\{x_{i,d}^k\}_{k=1}^K$) are sampled in a similar way as in the imputation task.
This objective naturally balances the exploration among features as well as the exploitation to find informative ones for the prediction task. For example, if feature $x_{i,d}$ is less explored or uninformative about the target, 
the first entropy term in Eq. \ref{eq: Acquisition Function Prediction} will be high, which encourages the algorithm to pick this unfamiliar data. However, using this term alone can result in selecting uninformative/noisy features. Thus, a counter-balance force for exploitation is needed, which is exactly the role of the second term. Unless $x_{i,d}$ together with $\theta$ can provide extra information about the target $\bm{y}_i$, the entropy in the second term with uninformative features will still be high. Thus, the two terms combined together encourage the model to select the less explored but informative features. The proposed objective is mainly targeted at the second requirement mentioned at the beginning of this section. However, its effectiveness depends heavily on the imputing quality of $x_{i,d}$. Thus, a natural way to satisfy both conditions is a combination of the two objectives:
\begin{equation}
    R_C(x_{i,d},\bm{X}_O)=(1-\alpha)R_I(x_{i,d},\bm{X}_O)+\alpha R_P(x_{i,d},\bm{X}_O),
    \label{eq: Acquisition Combined}
\end{equation}
where $\alpha$ controls which task the model focuses on. This objective also has an  information theoretic interpretation. In the appendix \ref{subsec: EDDI}, we show that when $\alpha=\frac{1}{2}$, this combined objective is equivalent to the mutual information between $\theta$ and the feature-target pair $(x_{i,d},\bm{y}_i)$. 
\section{Experiments}
\label{sec:exp}
We evaluate Icebreaker first on machine learning benchmark data sets from UCI \citep{Dua:2017} on both imputation and prediction tasks. We then evaluate it in two real-world applications: (a) movie rating imputation  using \textit{MovieLens} \citep{harper2016movielens}; (b) risk prediction in intensive care using MIMIC \citep{johnson2016mimic}. 

\begin{figure}[t]
    \centering
    \subfloat[c][Boston Housing Imputation 
    ]{\includegraphics[width=0.3\textwidth,height=4cm]{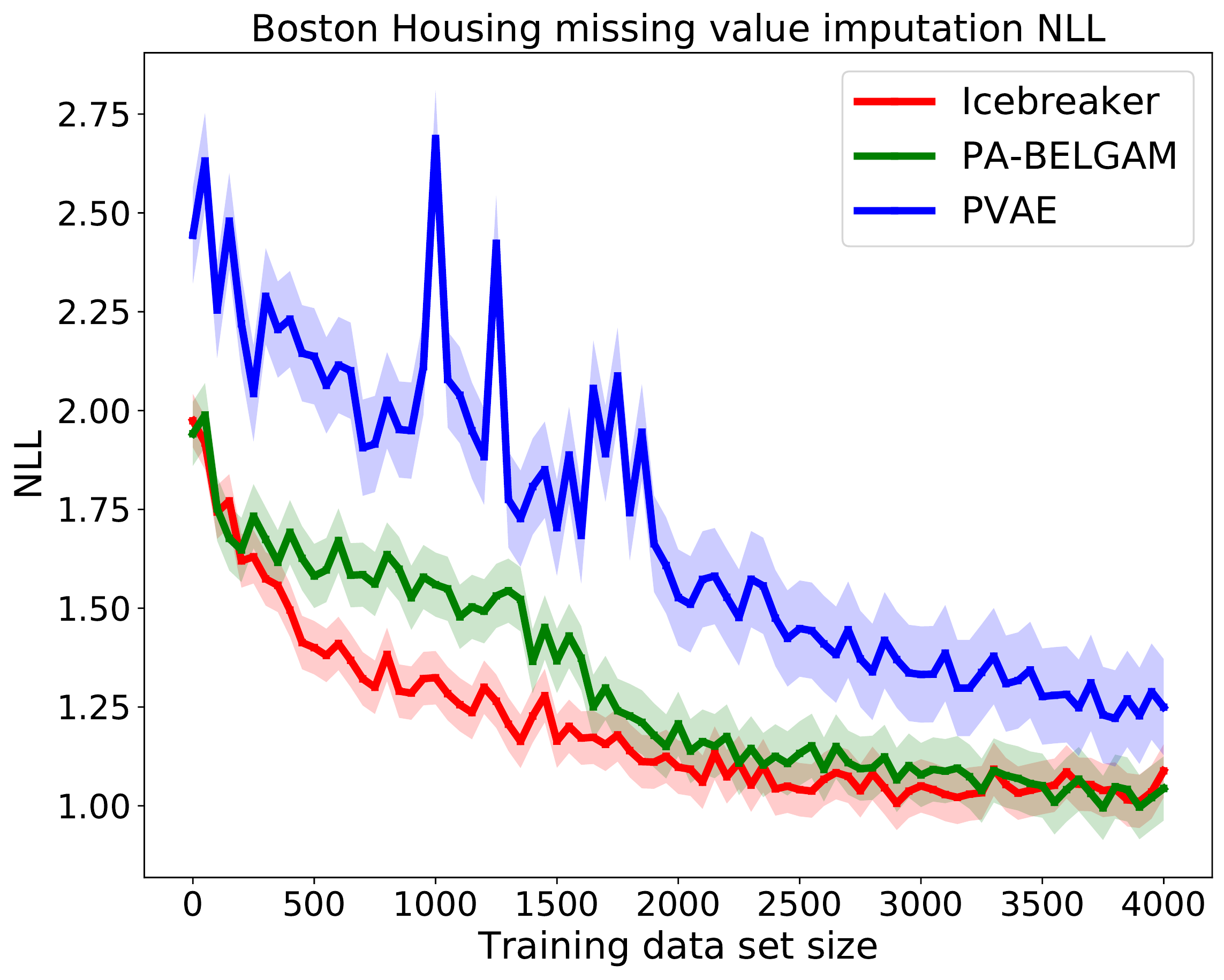}\label{fig: Boston Imputation}}\hfill
    \subfloat[c][Long tail selection pattern]{\raisebox{-0.2ex}{\includegraphics[width=0.3\textwidth,height=4.03cm]{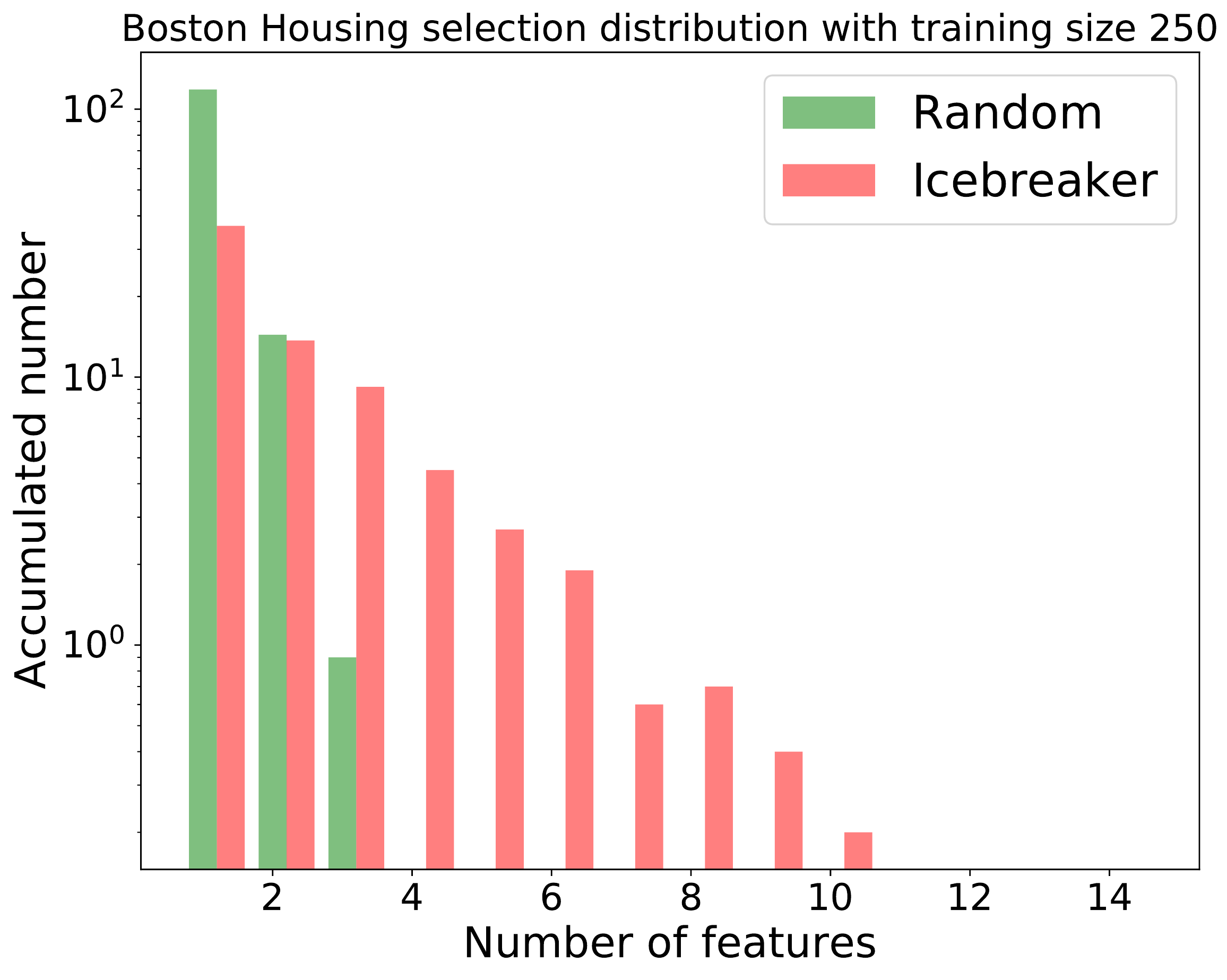}}\label{fig: Boston Imputation stat}}\hfill
    \subfloat[Boston Housing Active prediction]{\raisebox{0ex}{\includegraphics[width=0.36\textwidth,height=4cm]{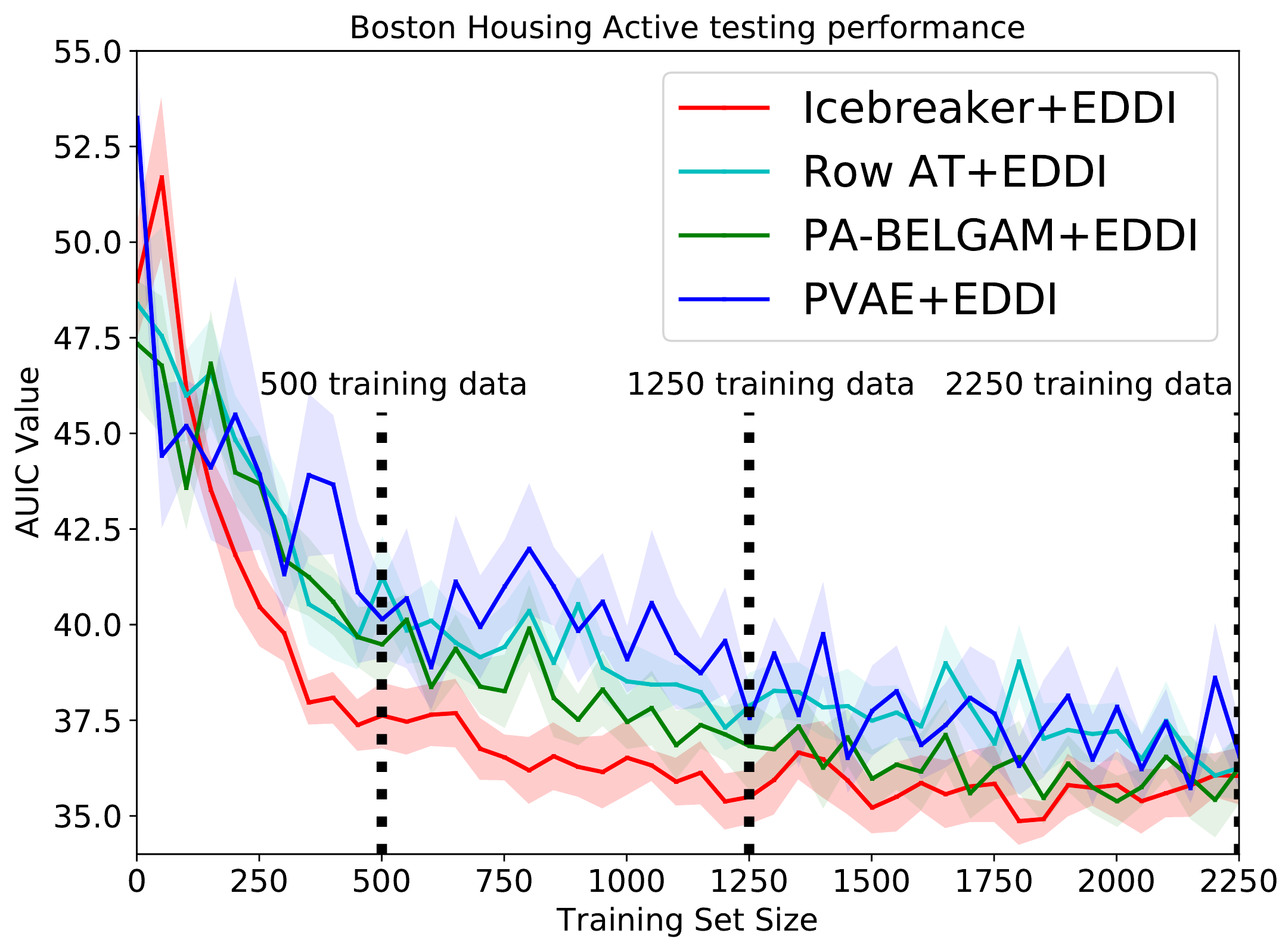}}\label{fig: Boston Prediction}}
    \caption{Boston Housing experimental results.  (a) The {NLL over the number of observed feature values. } (b) The distribution of the number of the feature for data instance  during the training time. (c) Performance of the active prediction task over  number of training data elements. The test time active prediction curves with the training data size indicated by black dash line are shown in Figure \ref{fig:Boston Test curve}}  
    \vspace{-10pt}
\end{figure}
\paragraph{Experiments Setup and evaluation.}
We compare the Icebreaker with random feature acquisition strategy for training where both  P-VAE \citep{ma2018eddi} and PA-BELGAM are used.  { For the imputation task, P-VAE already achieves excellent results in various data sets compared to  traditional methods \cite{ma2018eddi,nazabal2018handling}.} Additionally for the active prediction task, we compare the Icebreaker to the instance-wise active learning, denoted as \textit{Row AT}, where the data are assumed to be fully observed apart from the target. 

We evaluate the imputation performance by reporting \textit{negative log likelihood} (NLL) over the test target. For the active prediction task, we use EDDI \citep{ma2018eddi} to sequentially select features at test time.
We report the \textit{area under information curve} (AUIC) \citep{ma2018eddi} for the test set (See Figure \ref{fig:Boston Test curve} for example and appendix for details). A smaller value of AUIC indicates better overall active prediction performance. 
All experiments are averaged over 10 runs and their setting details are in the appendix.

\subsection{UCI Data Set}
\label{sec:UCI}

\paragraph{Imputation Task.} At each step of {Icebreaker} we select 50 feature elements from the pool. Figure \ref{fig: Boston Imputation} shows the {NLL on the test set} as the training set increases. {Icebreaker} outperforms random acquisition with both PA-BELGAM and P-VAE by a large {margin, especially at  early stages of training.} We also {see that} PA-BELGAM alone can be beneficial compared to {P-VAE} with small data set. {This is because P-VAE tends to over-fit, while PA-BELGAM leverages the model uncertainties.}

\begin{figure}[t]
    \centering
    \includegraphics[width=1\textwidth,height=3.8cm]{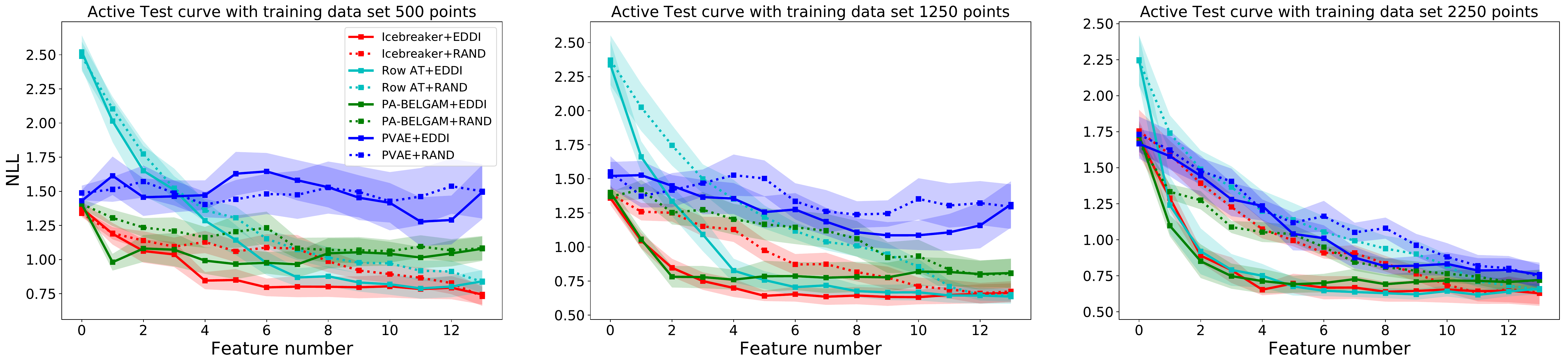}
    \caption{Evaluation of test time performance after exposure to different number of training data: (\textbf{Left}): 550 feature elements. (\textbf{Middle}):1250 feature elements (\textbf{Right}): 2250 feature elements. The x-axis indicates the number of features elements used for prediction. Legend indicates the methods used for training (Icebreaker, Row AT, etc.) and test time acquisition (EDDI, RAND)}
    \label{fig:Boston Test curve}
    
\end{figure}

We also analyze the feature selection pattern of the Icebreaker. We gather all the rows that have been queried with at least one feature during training acquisition and count how many features are queried for each. We repeat this for the first 5 acquisitions. Figure \ref{fig: Boston Imputation stat} shows the histogram of the number of features acquired for each data point. 
{The} random selection concentrates around 1 feature {per data instance}. However, the long-tailed {distribution of the number of selected features of} Icebreaker means it tends to exploit the correlations between features inside certain rows but simultaneously tries to spread its selection for more exploration.
We include the imputation results on other UCI data set in the Appendix. We find that Icebreaker consistently outperforms the baseline by a large margin.
\paragraph{Prediction Task.}
Figure \ref{fig: Boston Prediction} shows the AUIC curve as {the} amount of training data increases. The Icebreaker clearly achieves better results compared to all baselines (Also confirmed by Figure \ref{fig:Boston Test curve}). This shows that it not only has a more accurate prediction of the targets but also captures correlations between features and targets. Interestingly, the baseline \textit{Row AT} performs a little worse than PA-BELGAM. We argue this is because it is wasteful to query the whole row, especially with a fixed query budget. Thus, it will form a relatively small but complete data set. Again, the uncertainty of PA-BELGAM brings benefits compared to P-VAE with point estimated parameters.

We confirm our guess by plotting the active test NLL curve as \citep{ma2018eddi} in Figure \ref{fig:Boston Test curve}. At the early training stage (500 data points, {left panel in Figure \ref{fig:Boston Test curve}}), 
the performance of \textit{Row AT} is worse {in the test time} than others when few features are selected. {This is due to obtaining a complete observed datum is costly. With the budget of $500$ feature element, it can only select $50$ fully observed data instances. In contrast, Icebreaker has obtained 260 partially observed instances with different level of missingness.}
 As more features are selected during test, these issues are mitigated and the performance starts to improve.
Further evidence suggests that, as the training data grows, we can clearly observe a better prediction performance of \textit{Row AT} at {the} early test stage. We also include the evaluation of our method on other UCI data for active prediction in the appendix.


\begin{figure}[t]
    \centering
    \subfloat[Imputation NLL Curve]{\includegraphics[width=0.5\textwidth,height=4cm]{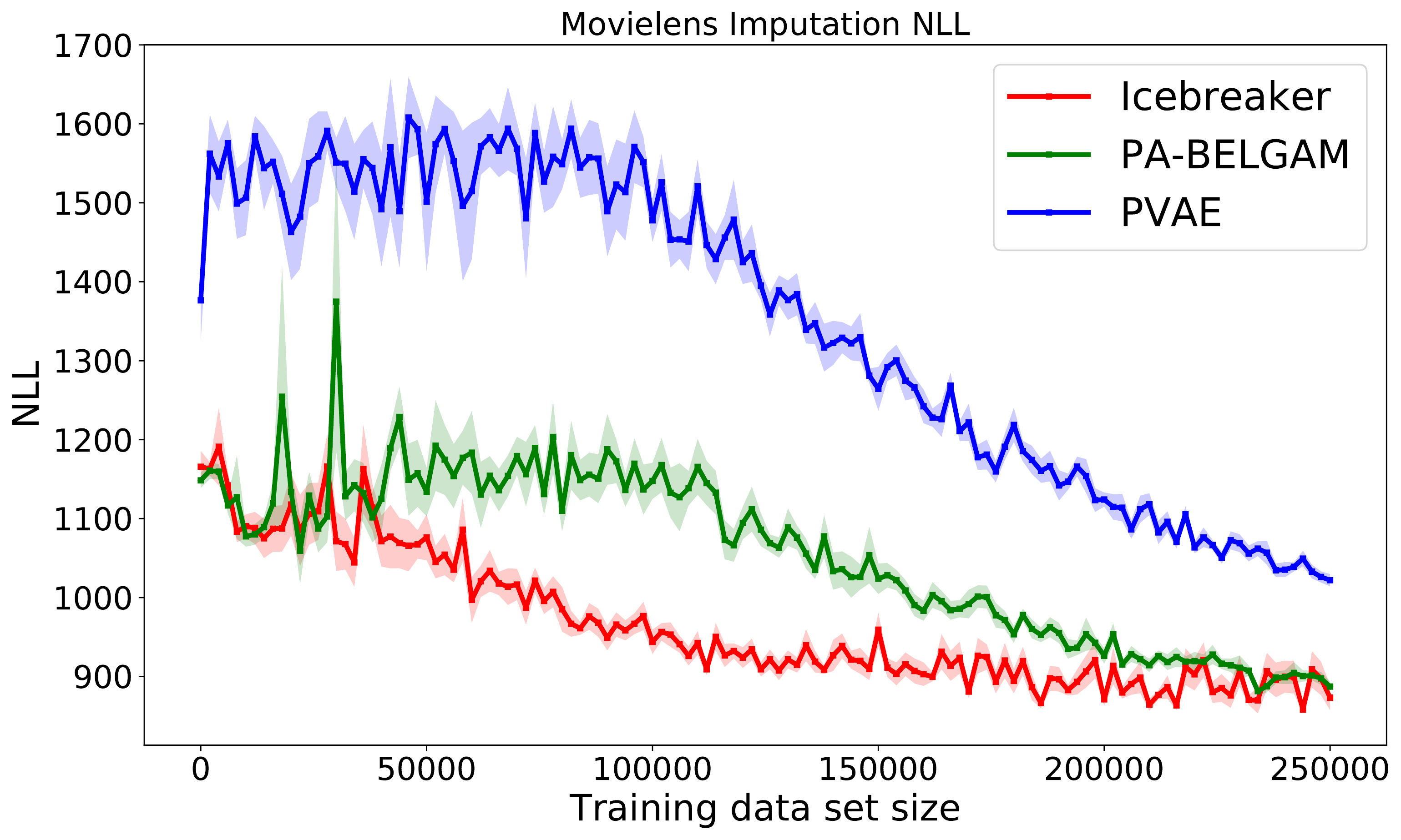}\label{fig: Movielens NLL}}
    \subfloat[Long-tailed selection]{\includegraphics[width=0.5\textwidth,height=4cm]{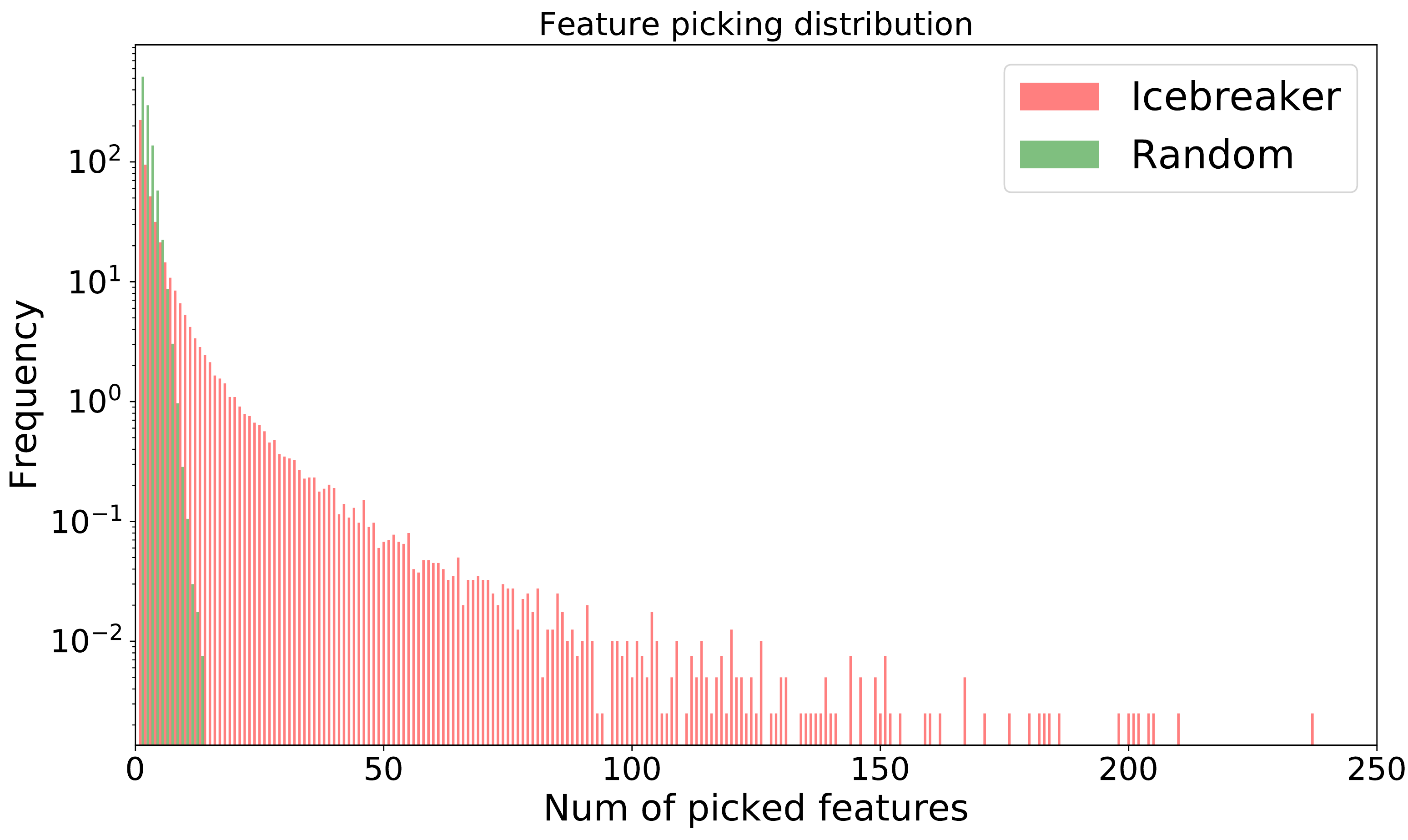}\label{fig: Movielens Long Tail}}
    \caption{Performance using MovieLens. Panel (a) shows the imputation NLL with number of observed movie ratings. Panel (b) shows the distribution of the number of features selected for each user).
    }
    \label{fig:movielens}
\end{figure}

\subsection{Recommender System using MovieLens}
\label{sec:MovieLens}
One common benchmark data set for recommender systems is \textit{MovieLens-1M} \citep{harper2016movielens}. 
P-VAE has obtained the state-of-the-art imputation performance after training with sufficient amount of data \citep{ma2018partial}. 
Figure \ref{fig: Movielens NLL} shows the performance on predicting unseen items in terms of NLL. Icebreaker shows that with minimum data, the model has learned to predict the unseen data. Given any small amount of data, Icebreaker obtains the best performance at the given query budget, followed by PA-BELGAM which outperforms P-VAE.
The selection pattern is similar to {the} UCI imputation, shown in Figure \ref{fig: Movielens Long Tail}. We argue this long tail selection is important especially when each row contains many features. 
The random selection tends to scatter the choices and is less likely to discover dependencies until {the} data set grows larger. But if there are many features per data instance, this accumulation will take very long time.
On the other hand, {the} long-tailed selection exploits the features inside certain rows to discover their dependencies and simultaneously tries to spread out the queries for exploration.

\subsection{Mortality Predicting using MIMIC}
\label{sec:MIMIC}
We apply the Icebreaker to a health-care application using the Medical Information Mart for Intensive Care (MIMIC III) data set \citep{johnson2016mimic}. {This is the largest real-world healthcare data set in terms of the patient number.} The goal is to predict the mortality based on the 17 {medical measurements}. The data is pre-processed following \cite{harutyunyan2017multitask} and balanced. Full details are available in the appendix \ref{subsec: MIMIC}. 

\begin{figure}[t]
    \centering
    \includegraphics[scale=0.14]{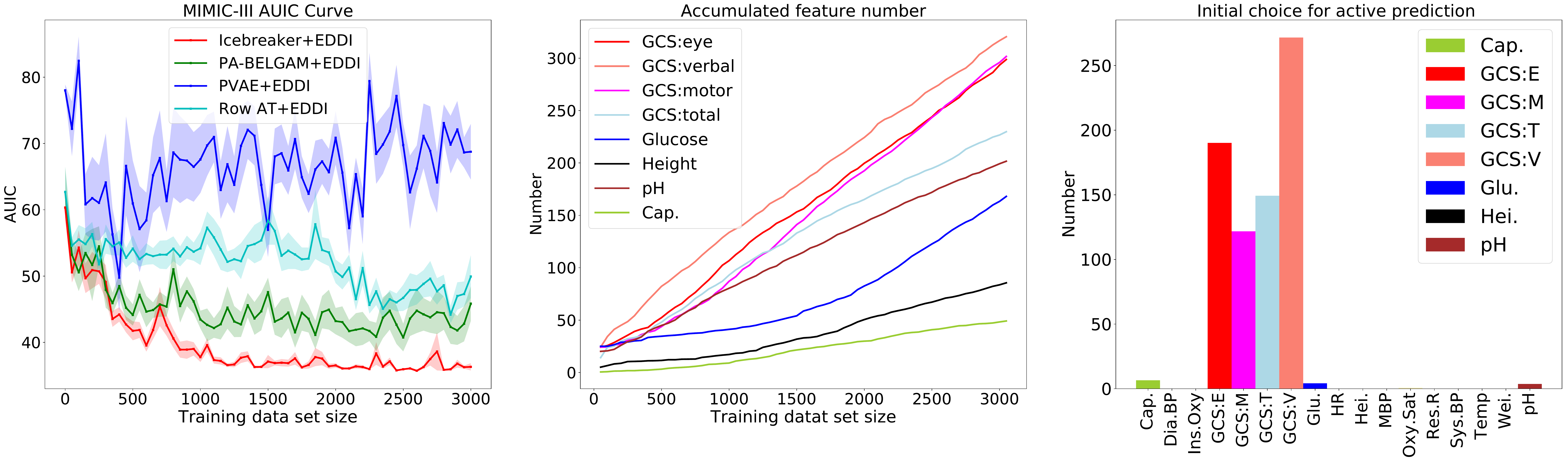}
    \caption{ {Performance of MIMIC experiments.} (\textbf{Left}) This figure shows the predictive AUIC curve as training data size increases. (\textbf{Middle}) The accumulated feature statistics as active selection progresses (\textbf{Right}) This indicates the histogram of initial choice during active prediction using EDDI.  
    }
    \vspace{-10pt}
    \label{fig:MIMIC}
\end{figure}

The left panel in Figure \ref{fig:MIMIC} shows that the Icebreaker outperforms the other baselines significantly with higher robustness (smaller std. error). Robustness is crucial in health-care settings as {the} cost of unstable model performance is high. Similarly, \textit{Row AT} performs more poorly until it accumulates sufficient data. {Note that without active training feature selection, PA-BELGAM performs better than P-VAE due to its ability to model uncertainty given this extremely noisy data set. }

To evaluate whether the proposed method can discover valuable information, we plot the accumulated feature number in the middle panel of Figure \ref{fig:MIMIC}. The x-axis indicates the total number of points in the training set and each point on the curve indicates {the} feature selection number in the training set. {We see that not only different features have been collected at different frequency, the curve of the accumulated feature such as Glucose is clearly non-linear as well. This indicates that the importance of different feature varies at different training phases. Icebreaker is establishing a sophisticated feature element acquisition scheme that no heuristic method can currently achieve. } The top 3 features are the \textit{Glasgow coma scale} (GCS). These features have been identified previously as being clinically important (e.g. by the IMPACT  model \citep{steyerberg2008predicting}. 
{\textit{Glucose} is also in the IMPACT set. It was not collected frequently in the early stage, but in the later training phase, more \textit{Glucose} has been selected. } 
{Compared to GCS, Glucose has a highly non-linear relationship with the patient outcome} \citep{popkes2019interpretable} (or refer to the appendix \ref{subsec: MIMIC}). {Icebreaker chooses more informative features with simpler relationship in the very early iteration. While the learning progresses, Icebreaker is able to identify these informative features with complex relationship to the target.} 
%
{Additionally, the missing rate for each feature in the entire data set differs. \textit{Capillary refill rate} (\textit{Cap.}) has more than $90\%$ data missing, much higher than \textit{Height}. Icebreaker is still able to pick the useful and rarely observed information, while only chooses a small percent of the irrelevant information during the test. }
On the right hand side of Figure \ref{fig:MIMIC}, we plot the histogram of the initial choices during test-time acquisition. \textit{GCS} are mostly selected {in the first step as it is the most informative feature}. 

\section{Related Work}

\paragraph{Data-wise Active Learning.}
The goal of active learning is to obtain optimal model performance with as fewer queries as possible \citep{mackay1992information,settles2012active,mccallumzy1998employing}, where only querying labels are associated with a cost. One category is based on decision theory \citep{roy2001toward}, where acquisition step is to minimize the loss defined by test tasks after making the query based on observed data. Indeed this is the optimal objective as it coincides perfectly with the goal of active learning. However, the evaluation of this objective can be expensive in practice \citep{kapoor2007selective,zhu2003combining}. Furthermore, it is task and model dependent, which makes if difficult to be applied in situations where test loss or distribution is unknwon in advance. Another category is based on information theory which is model and data set agnostic. Many previous active learning approaches are in this category. For example, \cite{tong2001support} proposed margin sampling based on heuristics, and \cite{cover2012elements,lindley1956measure} quantifies the uncertainty using KL divergence. Another well-known acquisition function is BALD \citep{houlsby2011bayesian} which is based on mutual information. Although our proposed acquisition for imputation is also based on mutual information, we emphasize that the original BALD objective is only applied to scenarios where the data set has fully observed inputs and target variables. In another words, those methods aim to only {select next data point to label while assuming that every feature of each data point is observed}. We call this approach \textit{instance-wise} selection. Obviously, these methods are not directly applicable to the settings considered here as {they assume that the only cost comes from acquiring labels}. {This is not the case in many real-world applications as any information acquisition is typically costly.}

\paragraph{Feature-wise Active Learning.}
Instead of only querying labels, the above active learning idea can be extended to query features, named as active feature acquisition (AFA). It makes sequential feature selections in order to improve model performance\citep{melville2004active,saar2009active,thahir2012efficient}. One category of these methods is based on estimating the expected improvement of the model, e.g. classification \citep{melville2005expected} and clustering \citep{vuintelligent}. However, this type of method is task specific and also expensive to compute. Another category of AFA is based on matrix completion. For example, in some cases, the observed features in a matrix are not enough to recover the missing entries. Thus, there are some active learning approaches proposed to query the most informative one for completion \citep{chakraborty2013active}. In \citep{sutherland2013active}, it queries the features for collaborative prediction using variational approximation of the posterior distribution, and \citep{ruchansky2015matrix} unifies the active querying and matrix completion. Instead of using low rank assumption for matrix completion, \citep{huang2018active} utilizes the supervised information from a classification model to complete the matrix and propose a variance based feature acquisition function. However, the above methods cannot be used for \textit{ice-start} problem as they either cannot handle prediction tasks (e.g. active matrix completion) or requires fully observed test input \citep{huang2018active}, i.e. they cannot solve \textit{active-prediction} problem. In addition, the feature acquisition function is also based on heuristics, which does not take \textit{aleoteric} uncertainty into account.
{Instead of actively selecting feature for training}, \cite{ma2018eddi} designs an information acquisition function to allow scalable active selection during test time. 
\cite{shim2018joint,janisch2017classification} use a RL-based decision making to actively select test features. However, these methods can only do active learning at test time and requires a large amount of training data. Our framework aims to principally combine the these two tasks. Namely, it can perform test-time active predictions but also can actively select training points from the very beginning, even in an ice-start situation. 
\paragraph{Cold-start problem} Another relevant problem to \textit{ice-start} is called \textit{cold-start} problem\citep{maltz1995pointing,schein2002methods}. The key difference between these two scenarios is that cold-start problem targets at the test time data scarcity after the model has been trained. Take recommender system as an example, the cold-start problem handles the scenario when there are new users incoming with no historical ratings, and the aim is to improve the recommendation quality for these new users. One common strategy is to utilise the meta data (e.g. user profiles, item category) to initialise the latent factors of users/items. \cite{xu2015ice,pandey2016resolving} initialized the users' latent factor by comparing the meta data of the new user with the old ones. On the other hand, \cite{houlsby2014cold} uses matrix factorization and information acquisition function to query the preference of the new user without using any meta data. However, one common assumption of these above methods is that the model is well-trained under sufficient amount of training data, which is not available under \textit{ice-start} scenario.
\section{Conclusion}
In this work, we introduce the ice-start problem where machine learning models are expected to be deployed where little or no training data has been collected. The costs of collecting new training datum apply at the level of feature elements. 
Icebreaker provides an information theoretical way to actively acquire  element-wise data for training and uses the minimum amount of data for  downstream test tasks like  \textit{imputation} and \textit{active prediction}.
Within the framework of \textit{Icebreaker}, we propose PA-BELGAM, a Bayesian deep latent Gaussian model together with a novel inference scheme that combines amortized inference and SGHMC. This enables  fast and accurate posterior inference. Furthermore, we propose two training time acquisition functions targeted at the \textit{imputation} and  \textit{active prediction} tasks. We evaluate \textit{Icebreaker} on several benchmark data sets including two real-world applications. Icebreaker consistently outperforms the baselines.  Possible future directions include taking the mixed-type variables into account and deploying it in a pure streaming environment. 
\bibliography{ref}
\bibliographystyle{abbrv}
\clearpage
\appendix
\section{Stochastic Gradient HMC}
Assume we want to draw samples $\theta\sim p(\theta|\bm{X}_O)$, the potential energy $U(\theta)$ is defined as (for our partial amortized inference algorithm, this is replaced with Eq. \ref{eq: SGHMC ELBO}) 
\begin{equation}
    U(\theta)=-\sum_{\bm{x}_I\in\bm{X}_O}{\log p(\bm{x}_i|\theta)}-\log p(\theta).
\end{equation}
Typically, we use the mini-batch to estimate this quantity, therefore, the stochastic estimate of $U(\theta)$ with {the} batch $\bm{X}_S$ can be written as 
\begin{equation}
    \tilde{U}(\theta,\bm{X}_S)=-\frac{|O|}{|S|}\sum_{\bm{x}_i\in\bm{X}_S}{\log p(\bm{x}_i|\theta)}-\log p(\theta),
\end{equation}
where $|S|$ and $|O|$ are the number of rows for $\bm{X}_S$ and $\bm{X}_O$ respectively.

The preconditioned SGHMC \citep{chen2016bridging} uses {the} diagonal Fisher information matrix as the adaptive pre-conditioner with moving average approximations\citep{li2016preconditioned}. Thus, {the} transition dynamics at time $t$ is {the} following :
\begin{equation}
    \begin{split}
        &\left.\begin{array}{lr}
             B=\frac{1}{2}\epsilon\\
             V_{t-1}=(1-\tau)V_{t-2}+\tau\nabla_{\theta}\tilde{U}(\theta)\cdot\nabla_{\theta}\tilde{U}(\theta)\\
             g_{t-1}=\frac{1}{\sqrt{\lambda+\sqrt{V_{t-1}}}}
        \end{array}\right\}\text{Preconditioning Computation}\\
        &\left.\begin{array}{lr}
             \bm{p}_t=(1-\epsilon\beta)\bm{p}_{t-1}-\epsilon g_{t-1}\nabla_{\theta}\tilde{U}(\theta)+\epsilon \frac{\partial g_{t-1}}{\partial \theta_{t-1}}+\sqrt{2\epsilon(\beta-B)}\eta\\
             \theta_{t}=\theta_{t-1}+\epsilon g_{t-1}\bm{p}_{t-1}
        \end{array}\right\}\text{SGHMC Updates}
    \end{split}
    \label{eq: SGHMC updates}
\end{equation}
where $\eta \sim \mathcal{N}(\bm{0},\bm{I})$ and $\epsilon$ is the step size. \cite{chen2016bridging} shows that the continuous-time dynamics of the above transitions can indeed preserve the stationary distribution $\pi(\theta)\propto \exp(-U(\theta))$.
In practice, the update equation of {the} preconditioning SGHMC Eq. \ref{eq: SGHMC updates} is closely related to Adam optimizer as discussed in \citep{chen2016bridging}. Intuitively, this can be regarded as a specially designed Adam with properly scaled Gaussian noise. Algorithm \ref{alg: Amortized + SGHMC} shows the procedure of {the} partial amortized inference.

\begin{algorithm}[H]
\SetAlgoLined
\SetNoFillComment
\SetKwInOut{Input}{input}

\Input{Data $\bm{X}_O$, step size $\epsilon$, friction $\beta$, thinning $\tau$, learning rate $\gamma$, initialized $\theta$, max sample size $N$}
\KwResult{Variational parameter $\phi$ and $\{\theta_n\}_{n=1}^N$ }
 Model and sampler initialization\;
 counter=0\;
 \While{not converged}{
  Sample minibatch $\bm{X}_S\in \bm{X}_O$\;
  Random masking with mask $\bm{m}$: $\tilde{\bm{X}}_S=\bm{X}_S\times \bm{m}$\;
  \tcc{Inference Network Update}
  Compute $\mathcal{L}(\tilde{\bm{X}}_S;\phi)$ using Eq.\ref{eq: Inference net ELBO}\;
  $q_\phi$: Optimize($\mathcal{L}(\tilde{\bm{X}}_S;\phi)$;Adam;$\gamma$)\;
  \tcc{SGHMC step}
  Compute $\tilde{U}(\theta)$ using Eq.\ref{eq: SGHMC ELBO} with proper scale\;
  $\theta$: Simulate dynamics Eq.\ref{eq: SGHMC updates}\;
  \tcc{Update the sample pool}
  \If{counter$=K\tau$, where $K$ is any positive integer}{$\{\theta_n\}=$Update($\{\theta_n\}$,$\theta$,N)\tcp*{Using FIFO procedure}}
  counter+=1\;
  
 }
 \caption{Amortized Inference + SGHMC}
 \label{alg: Amortized + SGHMC}
\end{algorithm}

\section{Conditional BELGAM}
\label{section: Conditional BELGAM}
We follow the similar notations as in main text, but we have additional target sets $\bm{Y}_O$ and $\bm{Y}^*$ in observed training and test data respectively. By similar derivations in \citep{sohn2015learning}, we have
\begin{equation}
    \log p(\bm{Y}_O|\bm{X}_O,\theta)\geq \sum_{i\in \bm{X}_O}{\left[
    \mathbb{E}_{q_{\phi}(\bm{z}_i|\bm{x}_i,y_i)}[\log p(y_i|\bm{z}_i,\theta)]-KL[q_{\phi}(\bm{z}_i|\bm{x}_i,y_i)||p(\bm{z}_i|\bm{x}_i)]
    \right]}
    \label{eq: CVAE}
\end{equation}
Note that the encoder proposed in \textit{BELGAM} can handle variable-sized inputs, thus, we can make further approximation $p(\bm{z}_i|\bm{x}_i)\approx q_{\phi}(\bm{z}_i|\bm{x}_i)$. We call the right hand side of Eq. \ref{eq: CVAE} as $\mathcal{L}_{conditional}(\bm{Y}_O;\phi)$. We should note that $\mathcal{L}_{conditional}(\bm{Y}_O;\phi)$ only focuses on prediction quality. On the contrary, successful active prediction, as discussed in main text, requires the model not only has a better target prediction but also capture the correlations between input features for sequential active decisions. Thus, in practice, we need to include the Eq. \ref{eq: Joint approximation} as well. Thus during each SGHMC step, Eq. \ref{eq: Joint approximation} is replaced with 
\begin{equation}
    \mathcal{L}(\{\bm{Y}_O,\bm{X}_O\};\phi)=\beta \mathcal{L}_{conditional}(\bm{Y}_O;\phi)+(1-\beta)\mathcal{L}_{joint}(\bm{X}_O;\phi)
    \label{eq: conditional SGHMC approximation}
\end{equation}
where $\beta$ controls which tasks the model focuses on. When $\beta=0.5$, we have 
\begin{equation}
    \log p(\{\bm{X}_O,\bm{Y}_O\},\theta)\geq \mathcal{L}(\{\bm{X}_O,\bm{Y}_O\};\theta)
\end{equation}
with equality holds when $q_\phi(\bm{z}_i|\bm{x}_i)=p(\bm{z}_i|\bm{x}_i)$ and $q_\phi(\bm{z}_i|\bm{x}_i,y_i)=p(\bm{z}_i|\bm{x}_i,y_i)$. In experiment, we choose $\beta=0.6$. We can also derive the equivalent form for Eq.\ref{eq: Inference net ELBO} using similar procedures for inference network update.

\section{Information acquisition}
\subsection{Theoretical results }
\label{subsec: EDDI}
\paragraph{Review: EDDI.}
For {the} active target prediction, model need to decide which feature should be queried for the purpose of predicting {the} target $\bm{Y}$ accurately in each test selection step. \cite{ma2018eddi} proposes a reward function for this test task inspired by Bayesian experimental design \citep{lindley1956measure,bernardo1979expected}. They propose to select the data point $x_{i,d}$ by maximizing:
\begin{equation}
    R(x_{i,d},\bm{X}_O)=\mathbb{E}_{x_{i,d}\sim p(x_{i,d}|\bm{X}_O)}\left[KL[p(\bm{y}_i|x_{i,d},\bm{X}_O)||p(\bm{y}_i|\bm{X}_O)]\right].
    \label{eq: EDDI}
\end{equation}
We find that this can be written as {the} mutual information between {the} target $\bm{y}_i$ and {the} candidate $x_{i,d}$:
\begin{equation}     R(x_{i,d},\bm{X}_O)=H[p(\bm{y}_i|\bm{X}_O)]-\mathbb{E}_{p(x_{i,d}|\bm{X}_O)}[H[p(\bm{y}_i|\bm{X}_O,x_{i,d})]].
    \label{eq: EDDI in Mutual Information }
\end{equation}
Thus, maximizing this quantity is equivalent to finding the most informative feature to {the} predictive target variable $\bm{y}_i$. However, this is not a suitable acquisition function in {the} training time as it is built on the assumption that the model is well trained and able to find the true informative features. Specifically, from Eq.\ref{eq: EDDI in Mutual Information }, it should be noted that $x_{i,d}$ is irrelevant to the first term. Thus, maximizing this objective is equivalent to minimizing the expected entropy after observing $x_{i,d}$, or conditional entropy $H(\bm{y}_i|x_{i,d})$. This objective purely encourages exploitation. 
For example, it can fail in the following scenario. In the beginning of training acquisition, the model may capture {the} wrong informative feature due to the small training data set. The exploitation nature of EDDI tends to pick this wrong feature over others in {the} following acquisitions and will be trapped into the sub-optimal strategy. 

\paragraph{EDDI for PA-BELGAM.}Next, we show that with a trained \textit{PA-BELGAM}, the above objective can be approximated efficiently. We assume the decoupled posterior $p(\theta,\bm{Z}|\bm{X}_o)\approx p(\theta|\bm{X}_O)p(\bm{Z}|\bm{X}_O)$ and conditionally independent features $p(\bm{x}_i|\bm{z}_i,\theta)=\prod_{d=1}^{|o_i|}p(x_{i,d}|\theta,\bm{z}_i)$. The EDDI rewards in Eq.\ref{eq: EDDI} can be rewritten by using KL chain rule:
\begin{equation}
    \begin{split}
        KL[p(\bm{y}_i|x_{i,d},\bm{X}_O)||p(\bm{y}_i|\bm{X}_O)]=&KL[p(\bm{y}_i,\bm{z}_i,\theta|x_{i,d},\bm{X}_O)||p(\bm{y}_i,\bm{z}_i,\theta|\bm{X}_O)]\\
        &-\mathbb{E}_{p(\bm{y}_i|x_{i,d},\bm{X}_O)}[KL[p(\bm{z}_i,\theta|\bm{y}_i,x_{i,d},\bm{X}_O)||p(\bm{z}_i,\theta|\bm{y}_i,\bm{X}_O)]].
    \end{split}
    \label{eq:EDDI step 0}
\end{equation}
The first term can be further approximated as 
\begin{equation}
    \begin{split}
        &KL[p(\bm{y}_i,\bm{z}_i,\theta|x_{i,d},\bm{X}_O)||p(\bm{y}_i,\bm{z}_i,\theta|\bm{X}_O)]=\begin{aligned}[t] &KL[p(\bm{z}_i,\theta|x_{i,d},\bm{X}_O)||p(\bm{z}_i,\theta|\bm{X}_O)]\\
        &+KL[p(\bm{y}_i|\bm{z}_i,\theta,x_{i,d},\bm{X}_O)||p(\bm{y}_i|\bm{z}_i,\theta,\bm{X}_O)]\end{aligned}\\
        &=KL[p(\bm{z}_i|x_{i,d},\bm{X}_O,\theta)||p(\bm{z}_i|\bm{X}_O,\theta)]+KL[p(\theta|x_{i,d},\bm{X}_O)||p(\theta|\bm{X}_O)]+KL[p(\bm{y}_i|\bm{z}_i,\theta)||p(\bm{y}_i|\bm{z}_i,\theta)]\\
        &=KL[p(\bm{z}_i|x_{i,d},\bm{X}_O)||p(\bm{z}_i|\bm{X}_O)].
    \end{split}
    \label{eq:EDDI step 1}
\end{equation}
where the last equality holds if we assume no posterior updates for $\theta$. This is a reasonable assumption because $x_{i,d}$ is only single data point adding into a much larger set $\bm{X}_O$. 
By using similar trick, we can show the second term in Eq.\ref{eq:EDDI step 0} is re-written as 
\begin{equation}
    \begin{split}
        &\mathbb{E}_{p(\bm{y}_i|x_{i,d},\bm{X}_O)}[KL[p(\bm{z}_i,\theta|\bm{y}_i,x_{i,d},\bm{X}_O)||p(\bm{z}_i,\theta|\bm{y}_i,\bm{X}_O)]]\\
        &=\mathbb{E}_{p(\bm{y}_i|x_{i,d},\bm{X}_O)}[KL[p(\bm{z}_i|\bm{y}_i,x_{i,d},\bm{X}_O)||p(\bm{z}_i|\bm{X}_O,\bm{y}_i)]+KL[p(\theta|\bm{y}_i,\bm{X}_O,x_{i,d})||p(\theta|\bm{X}_O,\bm{y}_i)]]\\
        &=\mathbb{E}_{p(\bm{y}_i|x_{i,d},\bm{X}_O)}[KL[p(\bm{z}_i|\bm{y}_i,x_{i,d},\bm{X}_O)||p(\bm{z}_i|\bm{X}_O,\bm{y}_i)]].
    \end{split}
    \label{eq:EDDI step 2}
\end{equation}
Then, we replace the posterior of $\bm{Z}$ with variational approximations $q_\phi$. Eq.\ref{eq: EDDI} can be aproximated as 
\begin{equation}
\begin{split}
     R(x_{i,d},\bm{X}_O)\approx &\mathbb{E}_{p(x_{i,d}|\bm{X}_O)}[KL[q_\phi(\bm{z}_i|x_{i,d},\bm{X}_O)||q_\phi(\bm{z}_i|\bm{X}_O)]]\\
     &-\mathbb{E}_{p(x_{i,d},\bm{y}_i|\bm{X}_O)}[KL[q_\phi(\bm{z}_i|\bm{y}_i,x_{i,d},\bm{X}_O)||q_\phi(\bm{z}_i|\bm{X}_O,\bm{y}_i)]].
\end{split}
\label{eq: EDDI Computable}
\end{equation}
This is exactly equivalent to the original form in \citep{ma2018eddi}. The only difference is the sampling stage for $x_{i,d}\sim p(x_{i,d}|\bm{X}_O)$ and $x_{i,d},\bm{y}_i\sim p(x_{i,d},\bm{y}_i|\bm{X}_O)$, where the $\theta$ samples are needed. 
\begin{equation}
    \begin{split}
        \bm{z}_i&\sim q_\phi(\bm{z}_i|\bm{X}_O)\\
        \theta&\sim p(\theta|\bm{X}_O) \;\;\; \text{using SGHMC}\\
        x_{i,d}&\sim p(x_{i,d}|\bm{z}_i,\theta)\\
        \bm{y}_i&\sim p(\bm{y}_i|\bm{z}_i,\theta)
    \end{split}
\end{equation}
\paragraph{Connections of Icebreaker acquisition function to mutual information}
We now show that the information acquisition function proposed in Eq.\ref{eq: Acquisition Combined} with $\alpha=\frac{1}{2}$ is equivalent to {the} mutual information between $\theta$ and {the} feature-target pair $(\bm{y}_i,x_{i,d})$.
\begin{equation}
\begin{split}
       R_{c}(x_{i,d},\bm{X}_O)&=\underbrace{\frac{1}{2}H[p(x_{i,d}|\bm{X}_O)]+\frac{1}{2}\mathbb{E}_{p(x_{i,d}|\bm{X}_O)}[H[p(\bm{y}_i|x_{i,d},\bm{X}_O)]]}_{\text{\circled{1}}}\\
       &\underbrace{-\frac{1}{2}\mathbb{E}_{p(\theta|\bm{X}_O)}[H[p(x_{i,d}|\theta,\bm{X}_O)]]-\frac{1}{2}\mathbb{E}_{p(\theta,x_{i,d}|\bm{X}_O)}[H[p(\bm{y}_i|\theta,x_{i,d},\bm{X}_O)]]}_{\text{\circled{2}}}.
\end{split}
\end{equation}
For \circled{1}, we have
\begin{equation}
    \begin{split}
        \text{\circled{1}}&=-\int{p(x_{i,d}|\bm{X}_O)\left[\log p(x_{i,d}|\bm{X}_O)+\int{p(\bm{y}_i|x_{i,d},\bm{X}_O)\log p(\bm{y}_i|x_{i,d},\bm{X}_O)d\bm{y}_i}\right]dx_{i,d}}\\
        &=-\int{p(x_{i,d}|\bm{X}_O)\int{p(\bm{y}_i|x_{i,d},\bm{X}_O)\log p(x_{i,d},\bm{y}_i|\bm{X}_O)}d\bm{y}_idx_{i,d}}\\
        &=H[p(\bm{y}_i,x_{i,d}|\bm{X}_O)].
    \end{split}
\end{equation}
For \circled{2}:
\begin{equation}
    \begin{split}
        \text{\circled{2}}&=\int{p(\theta,x_{i,d}|\bm{X}_O)\left[\log p(x_{i,d}|\theta,\bm{X}_O)+\int{p(\bm{y}_i|\theta,x_{i,d},\bm{X}_O)\log p(\bm{y}_i|\theta,x_{i,d},\bm{X}_O)d\bm{y}_i}\right]d\theta dx_{i,d}}\\
        &=\int{p(\theta,x_{i,d}|\bm{X}_O)\left[\int{p(\bm{y}_i|\theta,x_{i,d},\bm{X}_O)\log p(\bm{y}_i,x_{i,d}|\theta,\bm{X}_O)d\bm{y}_i}\right]d\theta dx_{i,d}}\\
        &=\int{p(\bm{y}_i,x_{i,d},\theta|\bm{X}_O)\log p(\bm{y}_i,x_{i,d}|\theta,\bm{X}_O)d\bm{y}_i dx_{i,d}d\theta}\\
        &=-\mathbb{E}_{p(\theta|\bm{X}_O)}[H[p(\bm{y}_i,x_{i,d}|\theta,\bm{X}_O)]].
    \end{split}
\end{equation}
Thus, the Eq.\ref{eq: Acquisition Combined} with $\alpha=\frac{1}{2}$ is written as 
\begin{equation}
    R_C(x_{i,d},\bm{X}_O)=\frac{1}{2}(H[p(\bm{y}_i,x_{i,d}|\bm{X}_O)]-\mathbb{E}_{p(\theta|\bm{X}_O)}[H[p(\bm{y}_i,x_{i,d}|\theta,\bm{X}_O)]])=\frac{1}{2}I(\theta,\{\bm{y}_i,x_{i,d}\}|\bm{X}_O).
\end{equation}

\section{Training details}
In this section, we give details about the experiment setup and {the} training acquisition.

\subsection{Training time acquisition}
We compute the Icebreaker acquisition functions (Eq.\ref{eq: Acquisition Combined} or Eq.\ref{eq: Acquisition Imputation}) for {the} entire pool set $\bm{X}_U$. During the selection ,we apply two heuristics. First, instead of picking the top K values, we first normalize their rewards $r_{id}$ with temperature $T$:
\begin{equation}
    w_{id}=\frac{\exp(r_{id}/T)}{\sum_{r_{id}}{\exp(r_{id}/T)}}
\end{equation}
Then, we sample $x_{i,d}$ according to their weights $w_{id}$. This is a common trick in {the} active learning techniques to encourage some exploration \citep{zhu2017generative,baram2004online}. When $T\rightarrow 0$, this sampling becomes {the} maximization. The second heuristic is to balance the selected feature number from the observed and new instances. Specifically, assume we need to select $K$ values from the pool, we use the above procedure to select $\frac{K}{2}$ from the rows that have been queried with at least one feature before and other $\frac{K}{2}$ from the rows that are completely new. This is to balance the proportion of exploiting the observed rows and exploring the new ones. For a fair comparison, the second heuristic is applied for all {the} baselines as well.
\subsection{Training hyperparameters}
\paragraph{UCI.} 
We split the whole data set into {the} training and test sets with proportion $80\%$ and $20\%$. In order to mimic that some features may not be available for query, we manually mask $20\%$ in {the} training set. For {the} imputation task, $40\%$ of the data in {the} test set are masked as {the} test target and the remaining $60\%$ are reserved as {the} test input. For {the} active prediction, we only mask the target variable in {the} test set as {the} test target. We also sample $2\%$ of the data instance as the pre-train data as the model has not learned anything in the beginning and the acquisition is the same as random. 

We use 5-dimensional latent variable $\bm{z}$ and the embedding for each feature $\bm{e}_d$ has 10 dimensions. $h(\cdot)$ is a neural network with 1 hidden layer of 20 units. The aggregation function $g(\cdot)$ is {the} summation. For {the} decoder, it has the structure $5-100-40-X$, where $X$ is the output dimensions. The data set is normalized with 0 mean and unit variance. We use $\alpha=1$ and $\alpha=0.4$ in Eq.\ref{eq: Acquisition Combined} for {the} imputation and {the} active prediction training time acquisition respectively. We use {the} learning rate 0.003 for {the} Adam optimizer and $\epsilon^2=0.0003$ for {the} SGHMC step size. We also use $\tau=0.99$, and $\epsilon\beta=0.1$ for {the} SGHMC hyperparameters. The model is trained with 1500 epochs and 100 mini-batch size. The first 750 epochs are used for {the} SGHMC burn-in and no $\theta$ samples are recorded.  At each training time acquisition, the model selects $25$ and $50$ values from the pool for {the} active prediction and {the} imputation respectively. 
\paragraph{MovieLens-1M.}
The \textit{MovieLens-1m} data set contains 1 million ratings for 3000 movies from 6000 users. Each rating is a categorical data ranging from 1 to 5. We follow the same data pre-processing procedure as \citep{houlsby2014cold} by selecting 1000 movies and 2000 users with the highest number of ratings. We follow {the} same settings as {the} UCI imputation with $0.5\%$ data as {the} pre-train and $20\%$ of the values in {the} test set as {the} targets. The model picks 5000 data points from {the} pool at {the} training acquisition followed by a model re-initialization to avoid local optima \citep{gal2017deep}.

The latent and feature embedding dimensions are 100 and 50 respectively. The decoder structure is the same as {the} UCI setting apart from the input and output dimensions. The learning rate and hyperparameters for Adam and SGHMC are the same as {the} UCI imputation. We train the model using 300 epochs with 100 batch size. Similarly we use half of the total epochs for {the} SGHMC burn-in. Each training acquisition selects 2000 values from the pool.
\paragraph{MIMIC III.}
The latent and feature embedding dimensions are the same as {the} UCI active prediction. The decoder structure is changed to $5-100-100-18$, where 18 is the data dimension of MIMIC III. The step size of SGHMC is changed to $\epsilon^2=0.0001$. The model is trained for 500 epochs with 100 batch size. The pre-train data set size is $0.5\%$ of the pool data. Each training acquisition selects 50 values from {the} pool set. The data normalization is the same as {the} UCI. 
\subsubsection{MIMIC III data set statistics}
\label{subsec: MIMIC}
\begin{figure}
    \centering
    \includegraphics[width=0.9\textwidth,height=7cm]{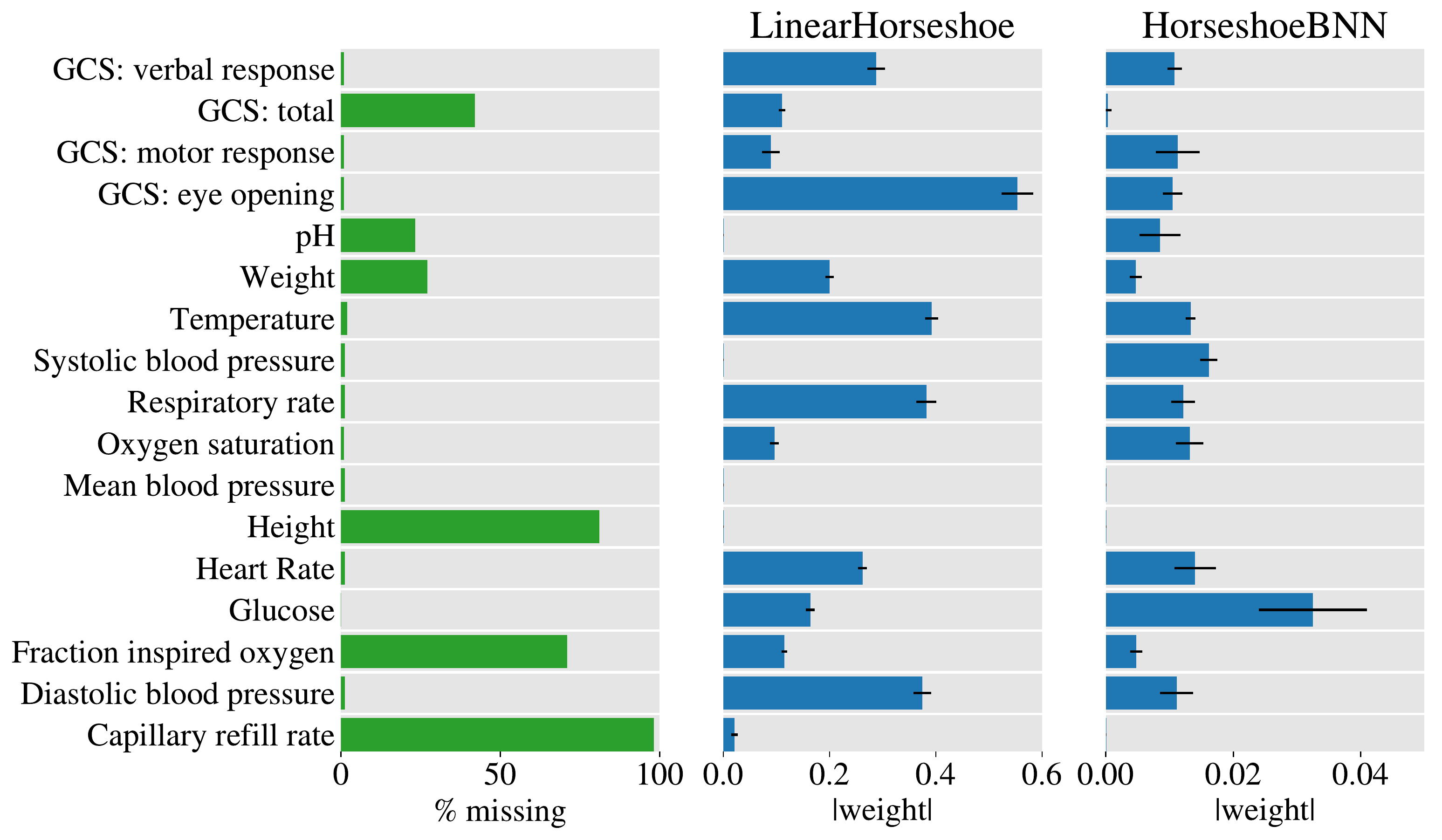}
    \caption{(Left) The missing proportion of each feature in MIMIC III. (Middle and Right) The norm of the weights. Zero feature weights indicate the corresponding feature is irrelevant to the target according to the model. This figure is directly taken from \cite{popkes2019interpretable}. }
    \label{fig: MIMIC Intepretable}
\end{figure}

MIMIC III data set after being processed by \citep{harutyunyan2017multitask} is extremely imbalanced, where around $88\%$ of the data has label $0$.  Thus, training with this data set will result in a lazy model that only outputs label 0. Typically additional pre-processing method for such data set is needed. In this project, we manually balance the data by taking an equal number of instances with label 0 and 1, which forms a new, balanced data set. We do the same for the test set as well. Figure \ref{fig: MIMIC Intepretable} (Left) shows the feature label and its missing proportion in MIMIC III. Table \ref{Table: Abbreviation} shows the acronym of each label.
From Figure \ref{fig: MIMIC Intepretable} (middle and right), we can observe there is a clear shift of importance for \textit{Glucose}. We hypothesize the reason is that the relationship of \textit{Glucose} and target is less linear and cannot be captured by the linear model. For the Icebreaker, when the training data set is small, it is easier to pick up simple relationships. Thus, \textit{Glucose} seems to be less relevant in the beginning. But as the data set grows, Icebreaker can capture non-linear dependencies and start to value the importance of \textit{Glucose}.  

\begin{table}[]
\centering
\begin{tabular}{|l|l|l|l|l}
\hline
\textbf{Acronym}           & \textbf{Label}           & \textbf{Abbrv.}           & \textbf{Label}        \\ \hline
\textit{\textbf{Cap.}}    & Capillary refill rate    & \textit{\textbf{Glu.}}    & Glucose               \\ \cline{1-4}
\textit{\textbf{Dia.BP}}  & Diastolic blood pressure & \textit{\textbf{HR}}      & Heart Rate            \\ \cline{1-4}
\textit{\textbf{Ins.Oxy}} & Fraction inspired oxygen & \textit{\textbf{Hei.}}    & Height                \\ \cline{1-4}
\textit{\textbf{GCS:E}}   & GCS: eye opening         & \textit{\textbf{MBP}}     & Mean blood pressure   \\ \cline{1-4}
\textit{\textbf{GCS:M}}   & GCS: motor response      & \textit{\textbf{Oxy.Sat}} & Oxygen saturation     \\ \cline{1-4}
\textit{\textbf{GCS:T}}   & GCS: total               & \textit{\textbf{Res.R}}   & Respiratory rate      \\ \cline{1-4}
\textit{\textbf{GCS:V}}   & GCS: verbal response     & \textit{\textbf{Temp}}    & Temperature           \\ \cline{1-4}
\textit{\textbf{Wei.}}    & Weight                   &                           &                       \\ \hline
\end{tabular}
\caption{Label and its Acronym}
\label{Table: Abbreviation}
\end{table}

\section{Additional UCI Results}
\begin{figure}
    \centering
    \subfloat[NLL Curve]{\includegraphics[scale=0.23]{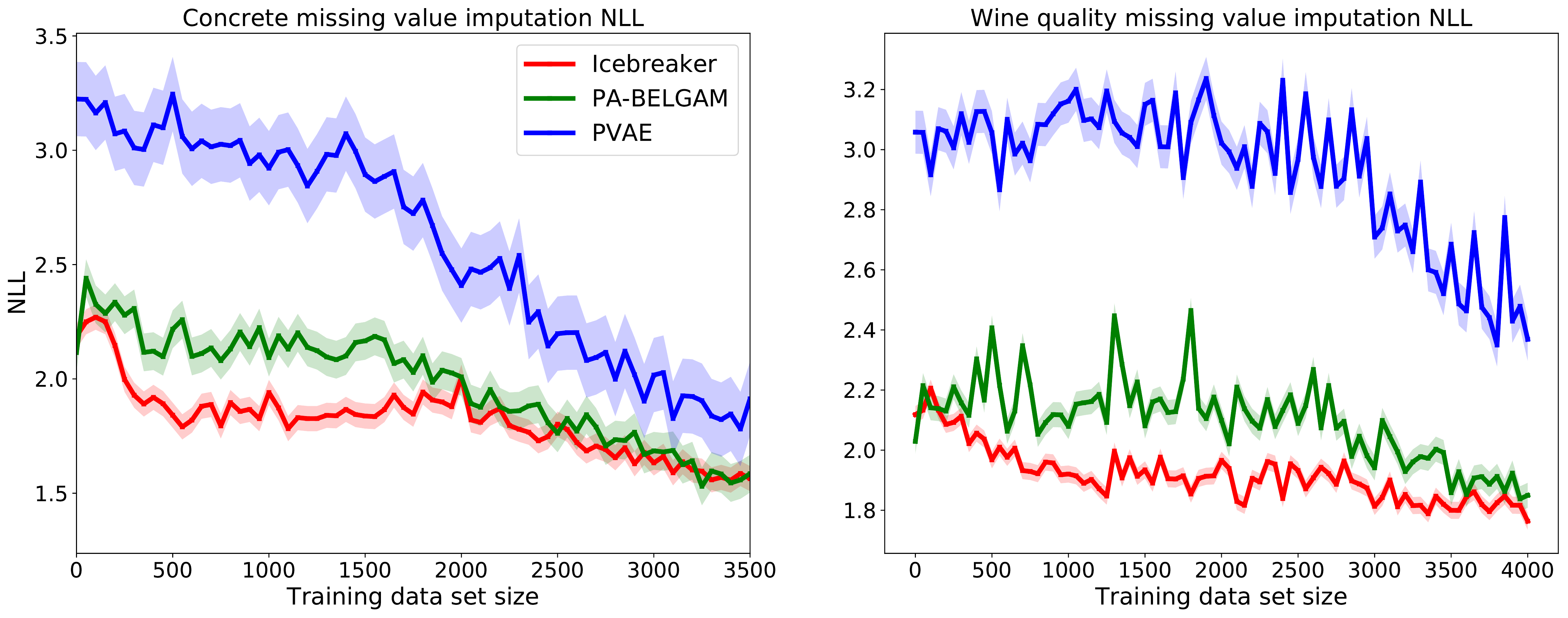}\label{fig: UCI Imputation Other}}\\
    \subfloat[Long Tail Selection]{\includegraphics[scale=0.23]{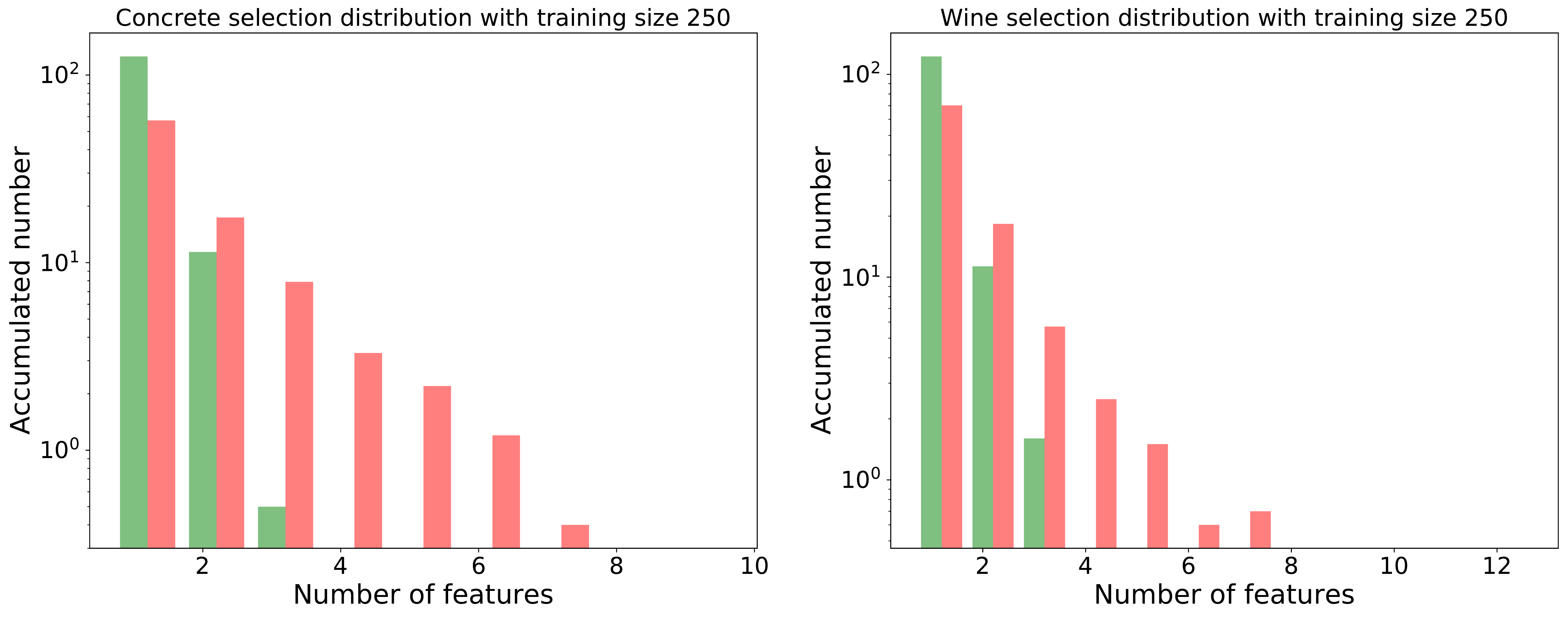}\label{fig: UCI Imputation Other Stat}}
    \caption{}
\end{figure}

For imputation task, we also evaluate the performance of Icebreaker on \textit{Concrete} and \textit{Wine quality} data sets. For the active prediction, we investigate its performance and feature selection strategy in a simple data set called \textit{Energy}. For all these experiments, we follow the experiment setup mentioned previously. 
\paragraph{Imputation.}
Figure \ref{fig: UCI Imputation Other} shows the imputation NLL curve as the data set grows. As expected, Icebreaker outperforms the baselines, especially when the training data size is small. Figure \ref{fig: UCI Imputation Other Stat} shows the selection pattern of Icebreaker, where we can observe a long-tailed strategy similar to \textit{Boston housing} and \textit{MovieLens-1m} imputation. 

\begin{figure}[!h]
    \centering
    \includegraphics[scale=0.27]{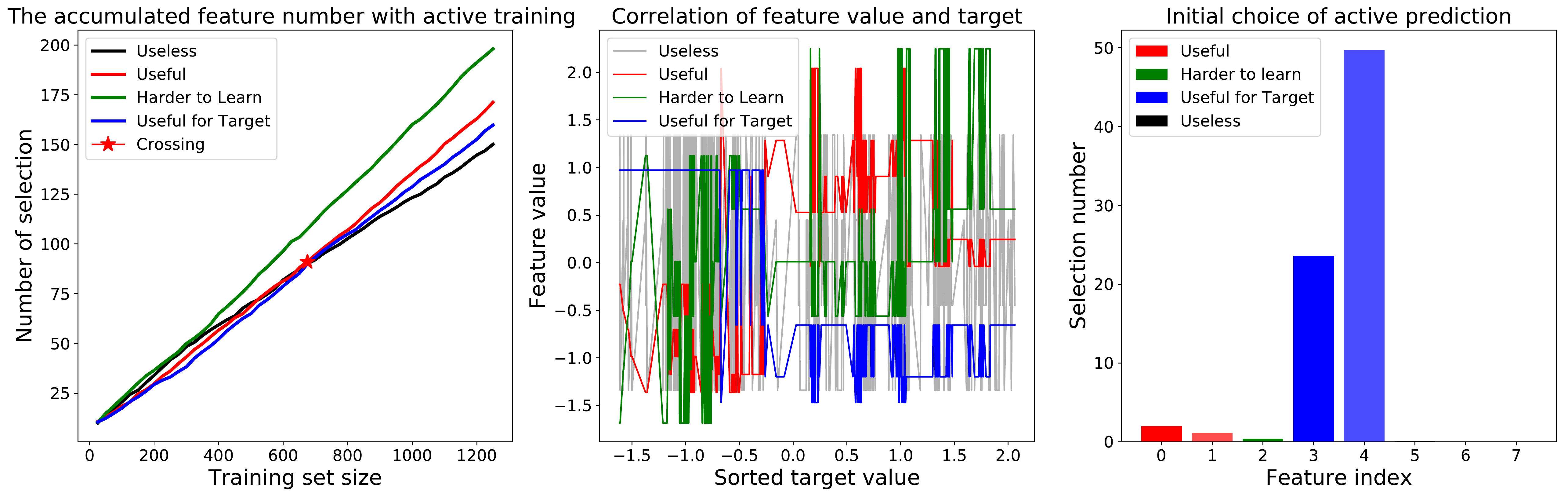}
    \caption{(Left) Accumulated feature number (Middle) Correlations between the features and target (Right) Initial choice at the test time acquisition.}
    \label{fig:UCI Prediction Stat 2}
\end{figure}

\paragraph{Active Prediction.}
\begin{figure}
    \centering
    \includegraphics[scale=0.4]{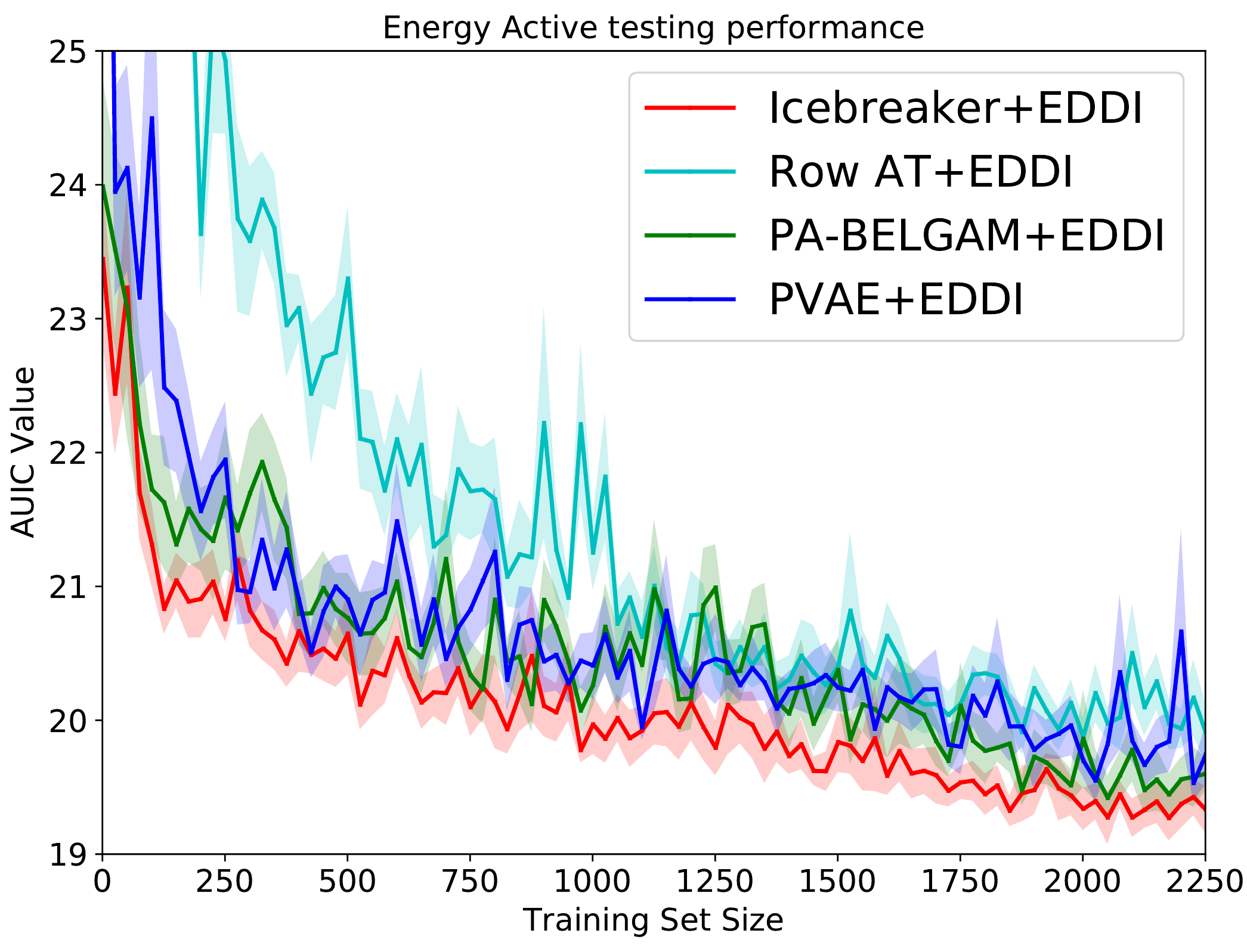}
    \caption{AUIC curve w.r.t. \textit{Energy} data set size}
    \label{fig: UCI Prediction 2}
\end{figure}
Figure \ref{fig: UCI Prediction 2} shows the AUIC curve as training data set size grows. We can see Icebreaker still outperforms the other 3 baselines by a small margin. The possible reason is that the \textit{Energy} data set is a very simple data set with clear informative variables. We choose this set for the purpose of diagnosing the selection strategy of Icebreaker rather than achieving significant improvement over others. To investigate the strategy of Icebreaker, we group the features in \textit{Energy} data set into 4 groups: \textcolor{blue}{Useful for target}, \textcolor{red}{Useful}, \textcolor{OliveGreen}{Harder to learn} and Useless based on the middle panel in Figure \ref{fig:UCI Prediction Stat 2}.

The x-axis in the middle panel of Figure \ref{fig:UCI Prediction Stat 2} represents the sorted target value from low to high. Y-axis indicates the feature values corresponding to the target. We can see for the \textcolor{blue}{blue} line, it has a clear boundary at target value $-0.3$ and the oscillation of this line is not large compared to others. Thus, it acts as an indicator feature to separate the small and large target values, which is the most useful one for predicting target. As for the \textcolor{red}{red} curve, it still has a relatively clear boundary but the large oscillation indicates this is not a robust feature for the prediction. So we refer to it as "\textcolor{red}{Useful}". Similar for \textcolor{OliveGreen}{green} curve, its boundary is less clear and it has a even larger oscillation. Therefore, we call it "\textcolor{OliveGreen}{harder to learn}". As for the black curve, it acts as the pure noise and has no clear correlations to target variables. We classify this as "Useless" features. 

From the left panel in Figure \ref{fig:UCI Prediction Stat 2}, initially, "\textcolor{OliveGreen}{Harder to Learn}" and "Useless" features are selected the most. This is because the objective Eq.\ref{eq: Acquisition Combined} encourages the model to find the informative but hard features. Due to the initially scarce training data, the model successfully figures out they are hard to learn but fails to identify which one is more informative. Thus, the selected elements for both features increases in a similar trend. With the data set growing, the model finds out the useless feature. Although it is hard to learn, the model still reduces its query frequency. As for the other two useful features, the model starts to select more of them after 800 data points. 

The right panel in Figure \ref{fig:UCI Prediction Stat 2} shows the initial choice made by the model during the active prediction. There are actually two features that can be classified as '\textcolor{blue}{Useful for target}'. But we only plot one of them in the left and middle panel of Figure \ref{fig:UCI Prediction Stat 2} for clarity. The other one is plotted in the right panel of Figure \ref{fig:UCI Prediction Stat 2} with light \textcolor{blue}{blue}. It is the same for '\textcolor{red}{Useful}' features.

It is expected that the '\textcolor{blue}{Useful for target}' feature is regarded as the most important one by the model though they are not selected the most in training. '\textcolor{red}{Useful}' and '\textcolor{OliveGreen}{Harder to Learn}' features are also selected with the number decreasing according to their importance. As expected, the 'Useless' features are not selected at all. Thus, the Icebreaker can indeed discover the important features and select the hard ones among them.  

\end{document}